\documentclass[sigconf]{acmart}

\AtBeginDocument{%
  \providecommand\BibTeX{{%
    \normalfont B\kern-0.5em{\scshape i\kern-0.25em b}\kern-0.8em\TeX}}}
\usepackage{tabularx}
\usepackage{bm}
\usepackage{algorithm}
\usepackage{algorithmic}
\copyrightyear{2022}
\acmYear{2022}
\setcopyright{acmcopyright}\acmConference[MM '22]{Proceedings of the 30th ACM International Conference on Multimedia}{October 10--14, 2022}{Lisboa, Portugal}
\acmBooktitle{Proceedings of the 30th ACM International Conference on Multimedia (MM '22), October 10--14, 2022, Lisboa, Portugal}
\acmPrice{15.00}
\acmDOI{10.1145/3503161.3547933}
\acmISBN{978-1-4503-9203-7/22/10}

\usepackage{bm}
\usepackage{multirow}
\usepackage{paralist}
\acmSubmissionID{761}

\begin{document}

\title{Few-shot Open-set Recognition Using Background as Unknowns}

\author{Nan Song}
\authornote{Nan Song is under the Joint PhD Program between Alibaba and Nanyang Technological University.}
\email{nan001@e.ntu.edu.sg}
\affiliation{%
  \institution{Nanyang Technological University \& Alibaba Group}
  \country{Singapore}
}

\author{Chi Zhang}
\email{johnczhang@tencent.com}
\affiliation{%
  \institution{Tencent}
  \country{China}
}

\author{Guosheng Lin}
\authornote{Corresponding author.}
\email{gslin@ntu.edu.sg}
\affiliation{%
  \institution{Nanyang Technological University}
  \country{Singapore}
}

\begin{abstract}
Few-shot open-set recognition aims to classify both seen and novel images given only limited training data of seen classes.
The challenge of this task is that the model is required not only to learn a discriminative classifier to classify the pre-defined classes with few training data but also to reject inputs from unseen classes that never appear at training time.
In this paper, we propose to solve the problem from two novel aspects. First, instead of learning the decision boundaries between seen classes, as is done in standard close-set classification, we reserve space for unseen classes, such that images located in these areas are recognized as the unseen classes. Second, to effectively learn such decision boundaries, we propose to utilize the background features from seen classes. As these background regions do not significantly contribute  to the decision of close-set classification, it is natural to use them as the pseudo unseen classes for classifier learning. Our extensive experiments show that our proposed method not only outperforms multiple baselines but also sets new state-of-the-art results on three popular benchmarks, namely tieredImageNet, miniImageNet, and Caltech-USCD Birds-200-2011 (CUB).

\end{abstract}



 \begin{teaserfigure}
	\centering
	\includegraphics[width=0.8\textwidth]{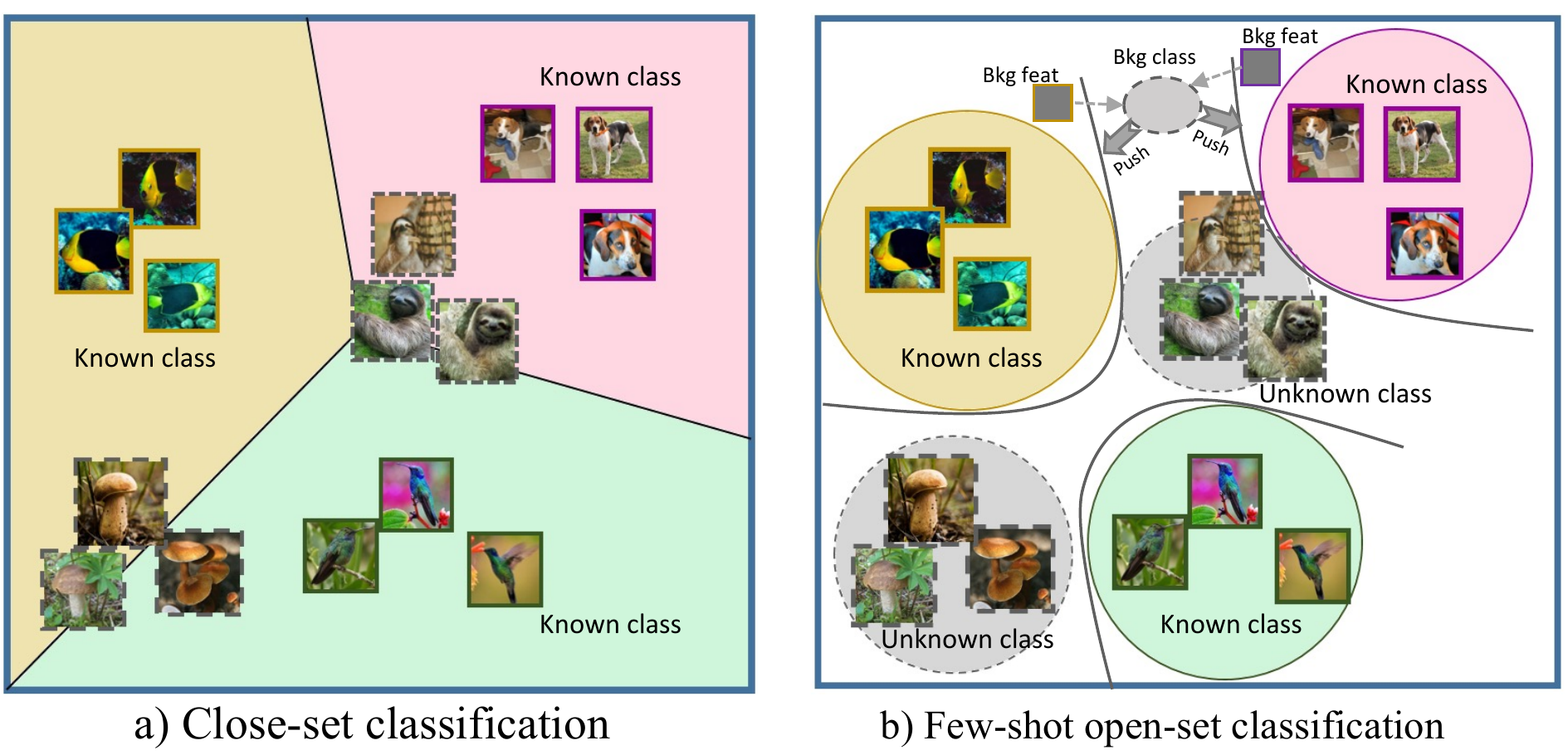}
	\caption{a) Close-set classification. Standard  classification classifies all  inputs to a set of pre-defined training classes.  b) Few-shot open-set classification task aims to classify images from seen and unseen classes given only limited training samples of seen classes. We propose to utilize the background regions at the training phase as the pseudo unseen classes to train the classifier that can directly classify seen and unseen images.
% 	Our proposed BKFAND reserve space for unknown class as background class and train the background class classifier with the background feature of known samples.
	}
	\label{fig:small}
\end{teaserfigure}

\maketitle

%%%%%%%%% BODY TEXT
\section{Introduction}
Deep Neural Networks have made great success in computer vision tasks, such as image recognition, image segmentation, text understanding, and object detection~\cite{zhang2021cyclesegnet,liu2020weakly,zhang2018efficient,daihong2022multi,li2022rigidflow}.   Deep models based on  Convolutional  Neural Network (CNN)~\cite{Alexnet,resnet,densenet,vgg} architectures have been proved efficient in excavating representative information from data in many fields. 

Supervised learning is one of the most dominant learning approaches for deep models, where the optimization of deep models can be achieved by feeding  a large amount of training data to the models and utilizing their labels as the training supervision. 
However, supervised learning limits the  model application to the pre-defined training classes, which is often described as the close-set learning task.
It is problematic to directly deploy these models to the real world, as a variety of novel categories that are out of the training classes may exist.

As it is impossible to enumerate all the real-world classes for training, the learned model has to recognize inputs from new classes as one of the known classes, as illustrated in Fig.~\ref{fig:small} (a), which hampers the applications of deep models in the real world. 

 Open-set recognition (OSR) is proposed, as a promising direction, to solve this kind of unknown class rejection problem. It aims to accurately classify the known classes and meanwhile detect the unknown classes.
 Many OSR methods make use of  the large amount of training data to estimate the distributions of training classes, such that those out-of-distribution data can be recognized as unseen classes.
 
This has largely solved the open-set recognition problem in many tasks. However, there are still many application scenarios where these algorithms are not applicable.
For example, in long-tailed recognition, when the training data of some classes,~\emph{e.g.}, rare animals, is not fairly enough, it is difficult to depict the distributions of these classes. Moreover, for tasks that rely on limited data to make predictions, such as few-shot learning, face verification, and person re-identification, these distribution-based methods would also fail.

To handle the problems in these special settings, a more challenging task, namely, Few-shot Open-Set Recognition (FSOSR) is proposed~\cite{peeler}, which aims to classify inputs from both known classes and unknown classes with only  a few labeled training data of seen classes available.

The goal of FSOSR model is twofold: to identify discriminative information from limited training data, and to have the ability to recognize unseen samples. The well-studied few-shot learning literature has well addressed the former problem, hence the main challenge of FSOSR is to detect unseen samples without harming the classification accuracy, which is the focus of this paper.
The main difficulty is that the information from training data to use for obtaining discriminative classifiers of seen classes is already very limited. A question is then raised --- where can we find the information of unseen classes?
In this paper, we argue that the background regions provide valuable information that can be utilized to detect samples from novel classes.
In closed-set recognition, a CNN is optimized to identify the discriminative information in the training samples, with the help of activation functions. As a result, only the discriminative parts in the images are kept as the representations of the data for classification, and the backgrounds are therefore ignored. Since these ignored backgrounds do not belong to any of these existing classes, we  treat them as the pseudo unseen classes to learn a classifier that can classify both seen and unseen classes meanwhile.
Instead of learning the decision boundaries of only seen classes, we reserve the space for unseen classes with the representations from the background, and we can easily conduct open-set classification by detecting samples located in these areas. Our motivation is illustrated in Fig.~\ref{fig:small} (b).

More specifically, we first initialize a class-mean classifier, where
the prototypes,~\emph{i.e.}, the weight vectors corresponding to different classes, are obtained  by averaging embeddings of support samples from each class.
We then randomly initialize a set of prototypes that correspond to unseen classes.
Next, we use the background features as well as the features of support data to finetune the whole classifier. The labels of the background features are set as the nearest prototypes that correspond to unseen classes.

\begin{figure}[t]
	\centering
	\includegraphics[width=1\linewidth]{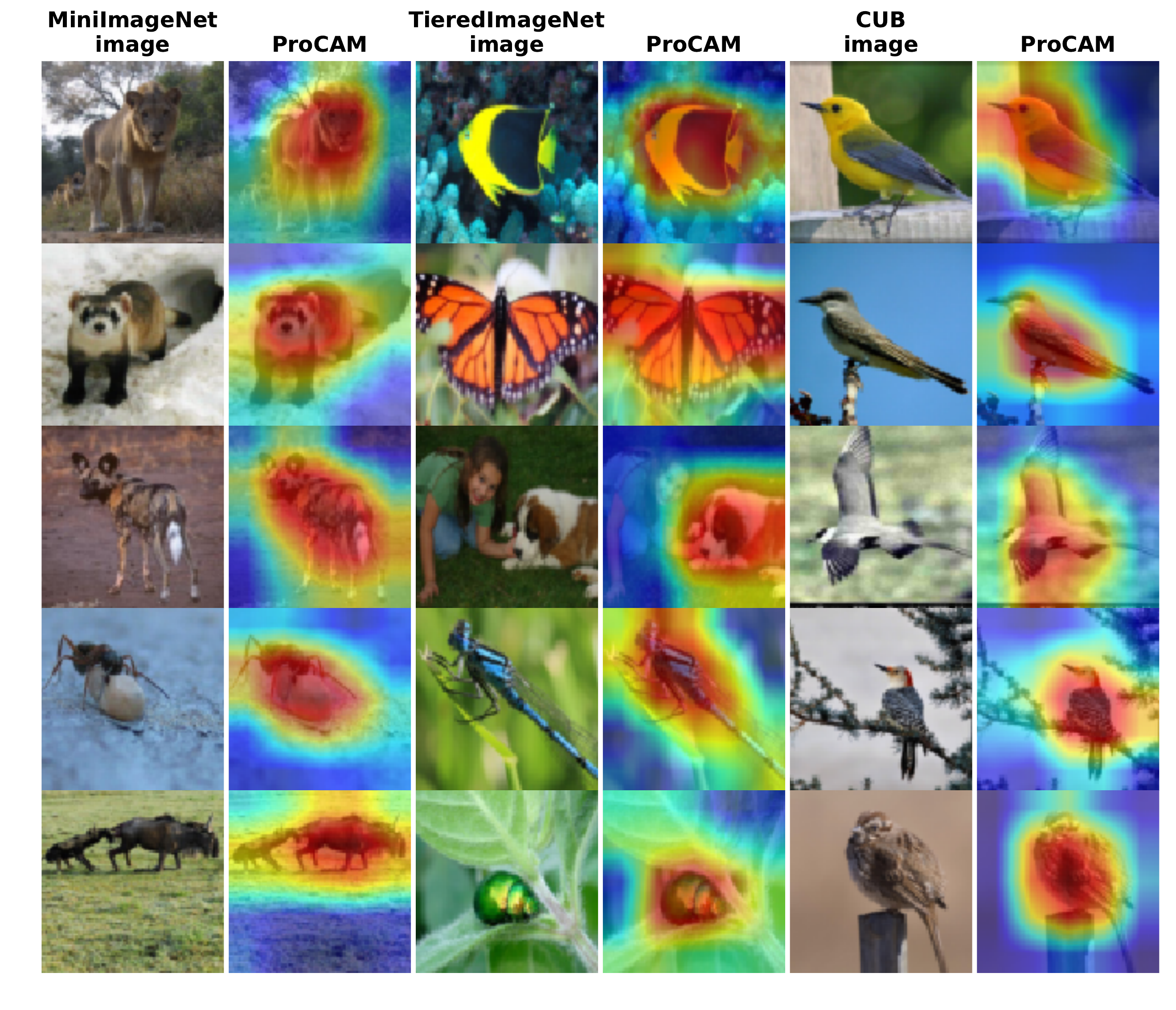}
	\caption{Our Progressive CAM (ProCAM) for the three benchmarks. The 1st column shows images from MiniImageNet and the ProCAM in the 2nd column highlights the foreground regions and accurately locates the background regions. The 3rd and 4th columns are from TieredImageNet and the last 2 columns are from CUB Birds.}
	\label{fig:cam_sample}
\vskip -2em
\end{figure}

 To accurately locate the background regions given limited training data, we further develop a Progressive Class Activation Mapping (ProCAM) module. Compared with the standard CAM~\cite{zhou2016learning}, the ProCAM  iteratively  mines the class-related regions in images, which can better separate the foreground and background areas in the low-show case as shown in Fig.~\ref{fig:cam_sample}. 
 We empirically show that our proposed ProCAM can not only generate better prototypes of unseen classes, but also regularize the learning of close-set classification.

Extensive experiments on multiple benchmark datasets have been conducted to verify the effectiveness of our proposed method. Our main contributions in this work can be summarized as follows.
\begin{itemize}
	\item We solve the few-shot open-set recognition task as a multi-class classification problem, where a set of additional classes is added as unseen classes. 
	\item To obtain the data used for training the classifier, we propose to utilize the background regions of images from the seen classes.
	\item We develop a Progressive CAM (ProCAM) module to iteratively seperate the class-related foreground regions and the background regions.
	\item Experiments on the tieredImageNet, CUB, and miniImageNet datasets show that our method significantly outperforms the baselines and sets new state-of-the-art performance with remarkable advantages.
\end{itemize}
\section{Related Work}

\begin{figure*}[t]
	\centering
	\includegraphics[width=0.92\linewidth]{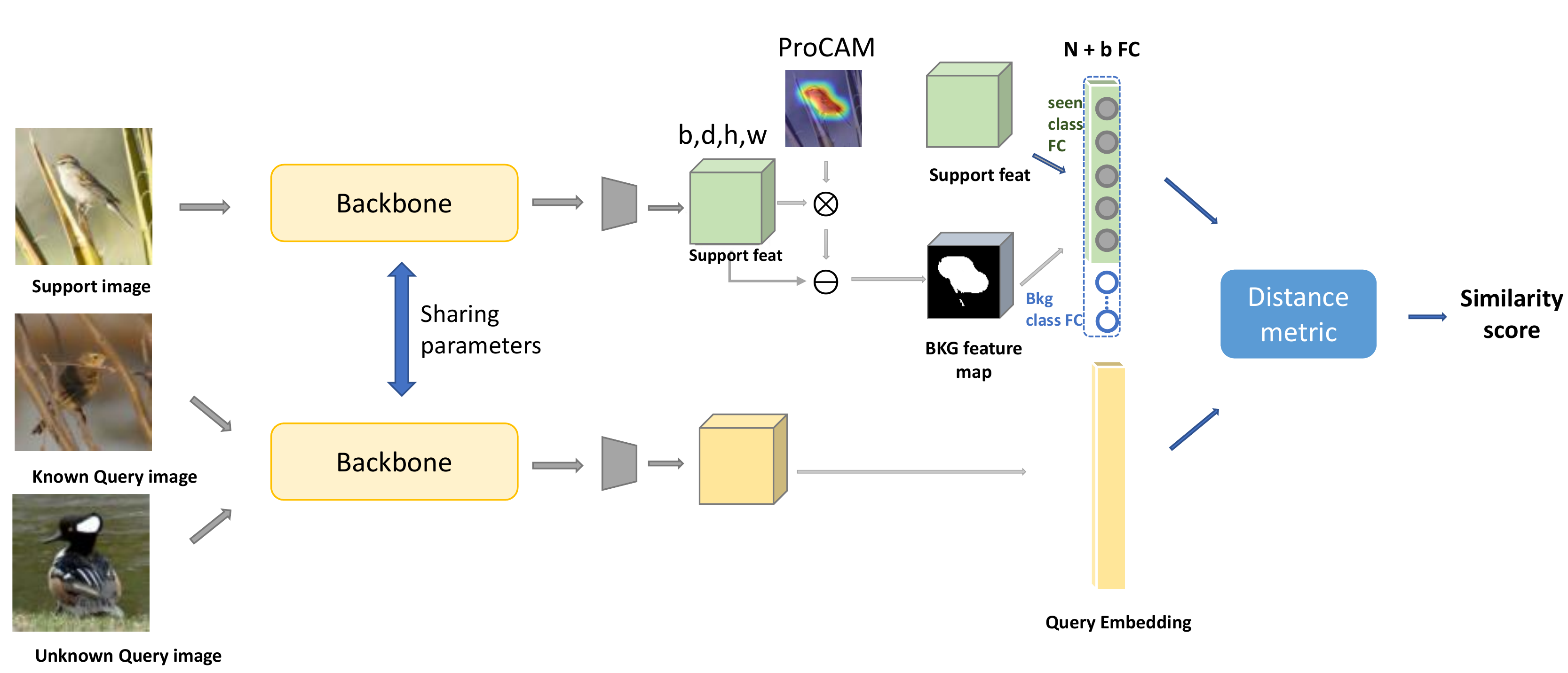}
	\caption{Our framework for few-shot open-set recognition. We introduce extra background classes to reserve space for unknown classes. Our background classifier uses background features as unknowns to optimize the classifiers.  We use a ProCAM module to collect multiple normalized class activation maps to find the foreground mask. Query embedding can be a known query or an unknown query, which is encoded with the same backbone. We use cosine similarity to obtain the class score.}
	\label{fig:whole}
\end{figure*}

\subsection{Few-shot learning}
Few-shot learning focuses on training models with insufficient training samples to gain the ability of classifying unseen images~\cite{ProtoNet,feat,RelationNet,Zhang_2020_CVPR,Zhang_2021_CVPR,Sun_2019_CVPR,Canfsl,zhang2021navigator,liu2021fewshot,zhang2019canet,zhang2020deepemdv2,liu2020crnet,chen2020compositional,pgnet,yang2022efficient}. There are great diversities shown up in the literature of few-shot learning. The two main streams in the few-shot learning literature are optimization-based methods~\cite{MAML,lee2019meta,LEO,Sun_2019_CVPR,omfsl,tasknorm} and metric-based methods~\cite{ProtoNet,MatchingNet,feat,Zhang_2020_CVPR,Canfsl,ye2019learning}. 

Optimization-based methods target at finding good initial parameters and adapting these parameters to new tasks effectively with low shot supports.
Ravi and Larochelle~\cite{omfsl} optimize the model parameters using LSTM in few-shot setting. Finn \textit{et al.}~\cite{MAML} proposed Model-Agnostic Meta-learning (MAML) where a meta-learner is designed to adapt parameters for crossed tasks. TaskNorm~\cite{tasknorm} proposed a normalization technique for meta-learning tasks to fuse different normalization values for efficient optimization.

Our model is more related to metric-based methods, which aim to distinguish different classes of instances with distance metrics in their feature space. MatchNet~\cite{MatchingNet} introduces cosine similarity to measure data distance and compute the similarity scores. ProtoNet~\cite{ProtoNet} proposes the negative L2 distance to measure the difference between the class prototypes and the query vector. DeepEMD~\cite{Zhang_2020_CVPR} applies Earth Mover's Distance (EMD) as their distance metric.
Recent methods transform the class prototypes to task-specific prototypes with task-adaptive weights~\cite{CTM} or transformer~\cite{feat}. 
The limitation of the existing FSL methods is that it only guarantees the good performance under the closed-set settings. The query set and the support set are in the same classes. FSOSR extends the FSL task with the unseen detection capability. We inject the additional background classes into conventional FSL methods and get a better detection capability.

\subsection{Open-set recognition}

Open-set recognition addresses the classification model with the detection capability of unseen samples~\cite{NN,OpenGAN,openmax,counterfactual,CrOSR,sun2020conditional,sun2020open,sun2021m2iosr}. It aims to reject unseen samples using the open-set classifier and classify the seen samples. The model cannot see the training samples for open-set data, and can only use the training data from the close-set.  Bendale \textit{et al.}~\cite{openmax} train a close-set classification network first and then endow the trained network with open-set discrimination using Extreme Value Theory. Gopenmax~\cite{Gopenmax} synthesizes fake open data during training. Neal \textit{et al.}~\cite{counterfactual} propose to generate counterfactual images for the unknown class and train an (N+1)-class classifier to recognize the N known classes and 1 unknown class. Zhou \textit{et al.}~\cite{zhou2021learning} also introduced the (N+1)-way classifier to solve the open-set recognition problem, but they synthesize fake open data using manifold mixup ~\cite{verma2019manifold} techniques, while we utilize image background.  Recently, GAN-based methods gain popularity~\cite{Gopenmax,c2ae,counterfactual,sun2020conditional,OpenGAN,wgan}. C2AE~\cite{c2ae} and CGDL~\cite{sun2020conditional} use all the training samples to train an autoencoder and detect unseen samples by the large reconstruction error.
Perera \textit{et al.}~\cite{perera2020generative} propose reconstructing the image along the channel dimension. The reconstruction of the unseen class samples is easy to fail which will lead to a lower classification probability for the unseen class.
The normal OSR methods all require rich training data to train the unseen sample detector~\cite{openmax, c2ae, sun2020conditional, perera2020generative, counterfactual}, which is hard to train in the FSL situation and getting poor performance.

\subsection{Few-shot open-set recognition}
FSOSR is recently proposed to address the open set recognition tasks in the few-shot inputs scenario~\cite{peeler,SnaTCHer}. PEELER~\cite{peeler} proposes the FSOSR task and solves the problem using the meta-learning framework and Mahalanobis distance to classify the known samples. They introduce an open-set loss using a negative entropy loss to maximize the entropy for seen classes. SnaTCHer~\cite{SnaTCHer} uses the pre-train model with the transformer from FEAT~\cite{feat} and needs to reuse the transformer to remap all the known and unknown query samples and a set of prototypes to a new space to calculate the distance. Our method is a metric-based FSOSR approach which is the same as PEELER and SnaTCHer, but we do not need the transformer module. PEELER~\cite{peeler} uses Mahalanobis distance. It needs to calculate the time-consuming covariance matrix. We follow MatchNet~\cite{MatchingNet} to use cosine distance which is simple and effective.

FSOSR is different from the Generalized zero-shot learning~\cite{liu2018generalized,ceGZSL} (GZSL). Although both do not have support samples for new classes, GZSL provides the semantic information for the target classes. FSOSR targets rejecting unseen new classes, the class information is unknown. As for GZSL, it is known information with class-level semantic description and target to classify target class samples to the correct classes.

\section{Problem Set-up}
FSOSR aims to design a machine learning algorithm that can reject unseen classes from only a few seen training examples and classify the seen classes samples correctly.\footnote{We interchangeably use seen and known, and unseen and unknown in this paper.} The typical setup for FSOSR follows episodic training and testing schema where batched tasks are sampled for training or testing. Specifically, an N-way K-shot FSOSR problem contains the support set and the query set for each episode. 

For support set $\mathcal{D}^{S}= \{ \bm{x}_{i}^{S}, y_{i}^{S} \}_{i=1}^{NK}$, where $\bm{x}_{i} \in \mathbf{X}_{S}$ and $y_{i} \in \mathbf{Y}_{S}$. The data from the support set $\mathbf{X}_{S}$ are from the classes seen with the label $\mathbf{Y}_{S}$. The query set for FSOSR is different from FSL, $\mathcal{D}^{Q} = \mathcal{D}^{K} \cup \mathcal{D}^{U}$, where $\mathcal{D}^{K}$ is the known query set and $\mathcal{D}^{U}$ is the unknown query set. The known query set are samples from the seen classes with the same label space $\mathbf{Y}_{S}$, $\mathcal{D}^{K} = \{ \bm{x}_{i}^{K}, y_{i}^{K}\}_{i=1}^{NQ}$, $\bm{x}_{i}^{K} \in \mathbf{X}_{K}$ and $y_{i}^{K} \in \mathbf{Y}_{S}$, $Q$ is the number of queries in each query class. The unknown query is sample from the unseen class with label space $\mathbf{Y}_{U}$. $\mathcal{D}^{U} = \{ \bm{x}_{i}^{U}, y_{i}^{U} \}_{i=1}^{UQ}$, $\bm{x}_{i}^{U} \in \mathbf{X}_{U}, y_{i}^{U} \in \mathbf{Y}_{U}$. There is no overlap for the seen labels and unseen $\mathbf{Y}_{S} \cap \mathbf{Y}_{U} = \phi$. 

An algorithm fits model $f(\bm{x})$ is defined as $f(\bm{x})=\mathbf{W}^{T}\Phi(\bm{x})$, where $\Phi(\cdot):\mathbb{R}^{D}\to\mathbb{R}^{d}$ is the embedding function and $\mathbf{W} \in \mathbb{R}^{d\times{N}}$ is the Fully Connected(FC) layer to classify seen classes. We denote the classifier for class $i$ as $\mathbf{w}_i:\mathbf{W}=[\mathbf{w}_1,\cdots, \mathbf{w}_{n}]$. 

\section{Method}
In this section, we introduce our framework for few-shot Open-set Recognition.
We first describe our extra background classifier method for clustering the unknown samples and separate the known classes in Section~\ref{section:ngeclass}. 
Then we present our proposed background feature as unknowns schema to train the extra background classifiers in Section~\ref{section:BFAND}.
The overview of the entire training pipeline is shown in Fig.~\ref{fig:whole}.

\subsection{Extra background classifier}
\label{section:ngeclass}
The purpose of Few-shot Open Set Recognition (OSR) is to separate unknown samples from known samples. The known samples need to do close-set classification in further steps. The difficulty lies in that unknown samples are incorrectly classified as known samples if the classifier only contains known classes. 
For a CNN classifier with $N$ classes, it will divide the embedding space into $N$ parts. The class boundaries are determined by the discriminative information in the classifiers. Unknown samples projecting to the same embedding space will fall into known classes space and classify wrongly with such kinds of classifiers. 
Therefore, we propose to reserve space for unknown classes, which is simple and straightforward by adding extra class classifiers. 
We get the hint from a special class in segmentation problems, the background class, which may contain a lot of unknown information compared to the foreground known classes. We name our extra class as background class for easy understanding. Let $\mathbf{W}_{bkg} \in \mathbb{R}^{d\times{N_{bkg}}} $ be the parameter matrix for the additional background classifier. It has the same number of feature channels $d$ as the known classes classifier $\mathbf{W}_k\in \mathbb{R}^{d\times{N_k}}$. $N_k$ is the number of known classes and $N_{bkg}$ is the number of background classes we pre-set to cluster possible unknown samples. The final FC of our network becomes $\mathbf{W} =[\mathbf{W}_k,\mathbf{W}_{bkg}] \in \mathbb{R}^{d\times (N_k+N_{bkg})}$. We employ the episodic training approach~\cite{ProtoNet} to train our model. Each mini-batch of training will mimic an FSOSR task, the classifier weight for each known class $i$ can be seen as the average embedding for each class $i$.

\begin{align} \label{eq:prototype}\textstyle
	\mathbf{w}_i=\frac{1}{K}
	{\sum_{\bm{x}_{j}^{S} \in \mathcal{X}_{i}^{S}}\Phi(\bm{x}_j)}.
\end{align}
We introduced the background features as unknowns method to train our background class classifier using the background features of the support data. The details are provided in Section~\ref{section:BFAND}

We use the cosine distance as our distance metric to measure the similarity between the query embedding $\Phi(\mathbf{X}_{Q})$ and the final FC of our network $\mathbf{W}$.
\begin{align} \label{eq:dist}\textstyle
	\textit{sim}\langle \Phi(\mathbf{X}_{Q}),\mathbf{W} \rangle =\big(\frac{W}{\|W\|_{2}}\big)^{\top}\big(\frac{\Phi(\mathbf{X})}{\|\Phi(\mathbf{X})\|_{2}}\big).
\end{align}
The highest similarity of $\Phi(\bm{x}_{i}^{Q})$ to the classifier weight $\mathbf{w}_y$  define the prediction of $\bm{x}$ belong to $y$.
The loss can be expressed as:
\begin{equation}
	\begin{split}
			\mathcal{L}(\bm{x},y)=\underbrace{\mathcal{L}_{CE} \left(f(\bm{x}^{K}),y^{K}\right)}_{\mathcal{L}_1}	+\lambda\underbrace{\,\mathcal{L}_{CE}\left(f(\bm{x}^{U}),\hat{y}\right)}_{\mathcal{L}_2},
		\end{split}
\label{eq:loss}
\end{equation}
where $\hat{y}=\textit{argmax}\{ \textit{sim}\langle\mathbf{W}_{bkg},\Phi(\bm{x}^U)\rangle\}+|Y_K|$ is the background class label with the maximum similarity the unknown query belongs to. The second term matches the unknown query to the reserved background classes space.

\begin{figure}[t]
	\centering
	\includegraphics[width=1\linewidth]{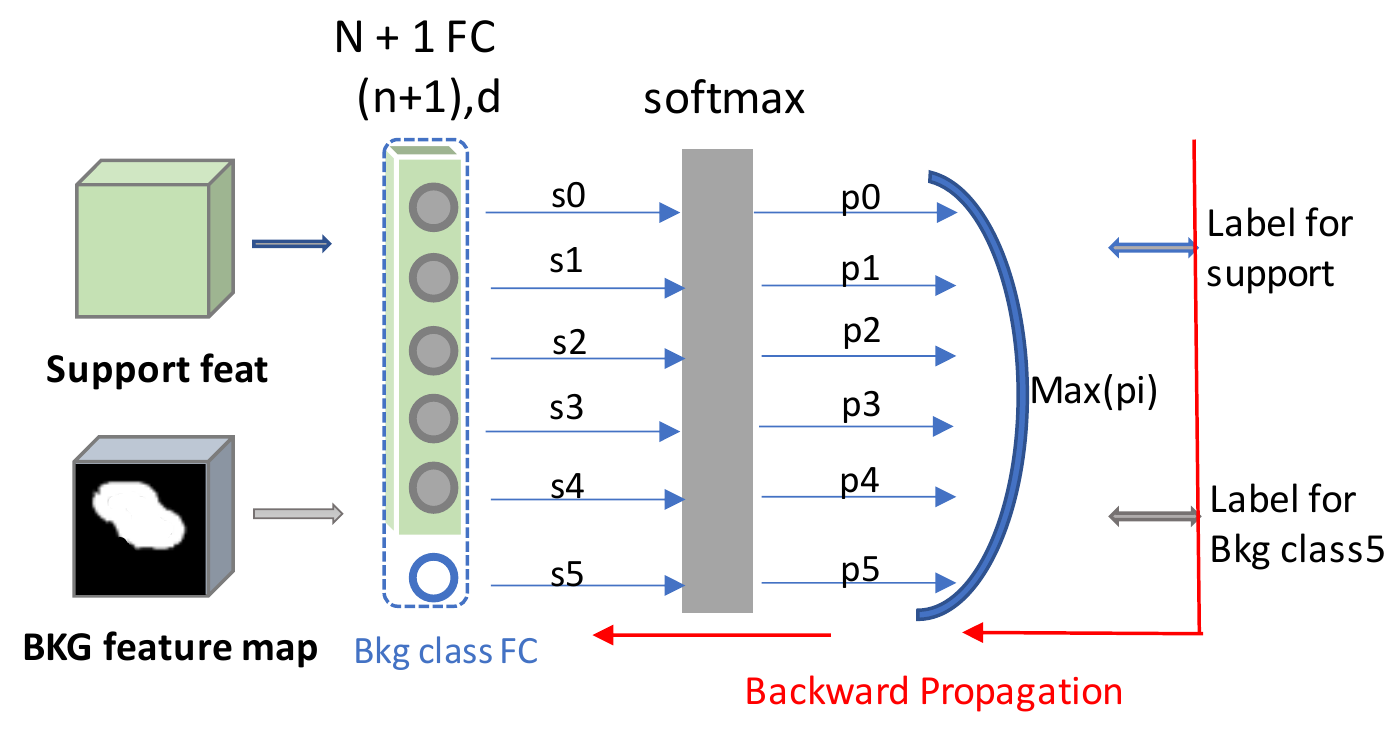}
	\caption{Our proposed method to finetune the background class using background features as unknowns. Here shows the details about background class training with $N_{bkg}=1$ as an example.}
	\label{fig:bkg_ft}
% \vskip -1.5em
\end{figure}

\subsection{Background feature as unknowns}
\label{section:BFAND}
FSOSR lacks open-set data to train the model. The existing OSR method normally uses GAN to generate pseudo-samples for open-set data, which is time-consuming and requires an extra branch of the generator network. We notice that the background of support data usually contains extra information and has a large difference from the foreground known class data. We can use background features as unknowns to train our extra background class and optimize the known classifier. Figure~\ref{fig:bkg_ft} shows the details of the classifier training. The training input is the support feature and the background feature of the support. The target label for the support feature is the known class label. We assign a background class label to the background feature. The backward propagation will only fine-tune the classifiers through the following loss.
\begin{equation}
	\begin{split}
			\mathcal{L}_b([\Phi(\bm{x}^{S}),\Phi_{bkg}])=\underbrace{\mathcal{L}_{CE} \left(\Phi(\bm{x}^{S}),y^{S}\right)}_{\mathcal{L}_3}	+\lambda_2\underbrace{\,\mathcal{L}_{CE}\left(\Phi_{bkg},\hat{y}\right)}_{\mathcal{L}_4},
		\end{split}
\label{eq:bfandloss}
\end{equation}
The background feature can be extracted from the class activation map of different known classes. We proposed a ProCAM module that utilized multiple normalized class activation maps to mine class-related regions and get better background features.

\subsection{ProCAM}
\label{section:normcam}
The ProCAM module is developed to apply CAM on the original support features multiple times to figure out the previously uncovered known class regions. The maps found multiple times can be superimposed to get more complete background area.
 This section introduces details on how to generate multiple class activation maps. A class activation map for a specific category will highlight the discriminative regions for that category. For a given input image with labeled class $c$, let $h_0(a,b)$ represent the activation of the image in spatial location $(a,b)$ in the last convolutional layer before average pooling. The CAM for class $c$ can be calculated through the following equation:
\begin{align} \label{eq:oricam}\textstyle
	M_c^0(x)=\sum_{all (a,b)}\mathbf{w}^{c} h_0(a,b),
\end{align}
where $\mathbf{w}^{c}$ is the classifier weight for class $c$. We scale $M_c^0(x)$ to $[0,1]$ using the Min-Max normalization. 
\begin{align} \label{eq:minimax_norm}\textstyle
	Norm(M_c^0(x))=\frac{M_c^0(x)-min(M_c^0(x))}{max(M_c^0(x))-min(M_c^0(x))},
\end{align}
The first background feature map is getting through $h_1=h_0\times(1-Norm(M_c^0(x)))$. We regard the first background as the input of the CAM equation again, the CAM function will highlight more discriminative regions to the class $c$. 
\begin{align} \label{eq:multiple_cam}\textstyle
	M_c^i(x)=\sum_{all (a,b)}\mathbf{w}^{c} h_i(a,b)),\newline
\end{align}
\begin{align} \label{eq:multiple_cam2}\textstyle
h_i(a,b)=h_{i-1}\times (1-\frac{M_c^{i-1}(x)-min(M_c^{i-1}(x))}{max(M_c^{i-1}(x))-min(M_c^{i-1}(x))}),
\end{align}
By repeating this operation for $\tau$ times can give us $\tau$ different CAMs $[M_c^0, M_c^1,\cdots,M_c^{\tau}(x)]$. We sum all these CAMs up and normalize them again to get the final ProCAM.
\begin{align} \label{eq:procam}\textstyle
	ProCAM=Norm(\sum_{i=1}^{\tau}M_c^{i}(x)).
\end{align}
The final background feature map $h_{bkg}=h_0(1-{ProCAM})$, where $h_0$ is the original feature map.
We improve ProCAM using softmax normalization to emphasize the high response regions and squeeze the low response regions to get a better mask for the foreground area. 
\begin{align} \label{eq:softmaxcam}\textstyle
    softmax(M_c(x))=\frac{\exp (M_c^i(x))}{\sum_{all(a,b)}\exp(M_c^i(a,b)) }.
\end{align}

\begin{algorithm}[t]
\caption{Background feature as unknowns.
$\tau$ is the number of multiple CAMs;
$y_K$ and $\hat y_{K}$ indicate the ground truth label and the network predictions for the known class, respectively;
$y_{bkg}$ and $\hat y_{bkg}$ indicates the ground truth label and the network predictions for the background class, respectively;
$\mathcal{L}(\cdot)$ is the cross-entropy loss function. 
$\lambda$ is the weight parameter
}
\begin{algorithmic}[1]
\REQUIRE Support datasets $\mathcal{D}_{train}^S$, pre-trained model $\mathcal{R}$, a randomly initialized FC for background classes $\mathbf{W}_{bkg}$.
\label{alg:PIL}
\ENSURE A trained FC for background classes $\mathbf{W}_{bkg}$ and known classes $\mathbf{W}_k$.
\WHILE{not done} 
\STATE $\mathbf{W}_{k} \leftarrow $ Learn FC layer upon $\mathcal{R}$ with $\mathcal{D}_{train}^S$
\STATE $h_0(x^S) \leftarrow$ Get the original feature map upon $\mathcal{R}$ with $\mathcal{D}_{train}^S$
\STATE $h_{tmp}(x^S)=h_0(x^S) \leftarrow$ Get a temporal copy of $h$
\STATE $cam\_all=[ ] \leftarrow$ An empty list to store all CAM 
    \FOR{ $i$ \textbf{in} range$(\tau)$}
        \STATE $M^i(x^S) \leftarrow$  Get CAM from $h_{tmp}(x^S)$ and $W_k$ with eq.~\ref{eq:oricam};
        \STATE $Norm(M^i(x^S)) \leftarrow$ Get normalized CAM using eq.~\ref{eq:minimax_norm};
        \STATE $cam\_all.append(Norm(M^i(x^S)))$
        \STATE $h_{tmp}(x^S)=h_{tmp}(1-Norm(M^i(x^S))) \leftarrow$ Update $h_{tmp}$ to bkg feature map 
    \ENDFOR
\STATE $ProCAM \leftarrow$ Get normalized multiple CAM using eq.~\ref{eq:procam};
\STATE $h_{bkg}=h_0(1-ProCAM)$
\STATE $\{\hat y_{K}, \hat y_{bkg}\} \leftarrow$ Make predictions for $\{h_0(x^S),h_{bkg}\}$ using $[\mathbf{W}_{k},\mathbf{W}_{bkg}]$ 
\STATE loss3 $\leftarrow$ Compute loss with $ \mathcal{L}_3(y_{K},\hat y_{K})$, 
\STATE loss4 $\leftarrow$ Compute loss with $ \mathcal{L}_4(y_{bkg},\hat y_{bkg})$, 
\STATE loss=loss3+$\lambda$ loss4 $\leftarrow$ get total loss, 
\STATE $\{\mathbf{W}_k',\mathbf{W}_{bkg}'\}\leftarrow$ Update classifier with SGD
\ENDWHILE
\end{algorithmic}
\end{algorithm}
\section{Experiment}

\begin{table*}[t]
\centering

\begin{tabular}{lcccccc}
\hline
\multicolumn{1}{l}{Model}   & meta learning & cam  &  ft\_all\_proto
&ProCAM& acc&auroc\\ \hline\hline
base &   &   &&&82.53  &78.54      \\ 
meta\_cos & \checkmark &   &&&84.59 &80.95              \\
meta\_cam & \checkmark &  \checkmark &&& 85.08      &82.41    \\ 
meta\_cam\_ft & \checkmark & \checkmark &\checkmark&&85.67  &82.45        \\ 
meta\_cam\_ft\_Norm & \checkmark & \checkmark &\checkmark&\checkmark&86.34  & 82.76      \\ 
\hline
\end{tabular}%

\caption{Ablation study on different components for our method on 5-shot TieredImageNet dataset}

\label{table:ablac}
\vskip -1em
\end{table*}
 
\subsection{Datasets}

\textbf{\emph{mini}ImageNet}~\cite{MatchingNet}: It is a popular benchmark for few-shot learning with 100 classes and 600 images in each class. ~\cite{peeler} proposed to use this dataset for the FSOSR task. Following the conventional splitting~\cite{MatchingNet} of meta-training(64), meta-validation(16), and meta-testing(20) for FSL, the few-shot open-set problem samples training episodes from the 64 meta-training classes, and testing episodes from the 20 meta-testing classes.

\begin{table}[t]
\centering

{%
\begin{tabular}{cccc}
\hline
 init. method & f.t. epochs  &  acc &auroc\\ \hline\hline
 random & epd10 &83.05  &76.15        \\ 
random & epd20 &\textbf{83.73} &77.44              \\
 random &  epd30 &83.60      &\textbf{77.48}    \\ 
 random &  epd50 &83.47      &77.28    \\ 
 random &  epd100 &83.26     &76.47    \\ 
\hline
 avg & epd10 &76.35  &73.71        \\ 
 avg & epd20 &76.49 &75.50               \\
 avg &  epd30 &\textbf{76.69}      &75.48    \\ 
 avg &  epd50 &76.65      &74.08   \\ 
 avg &  epd100 &76.36      &\textbf{76.48}    \\ 
\hline
 global & epd10 &\textbf{75.51}  &75.29         \\
 global & epd20 &74.86  &75.56       \\
  global & epd30 &75.00  &\textbf{75.90}        \\
   global & epd50 &74.04  &75.56         \\
global & epd100 &73.99  &75.52   \\\hline
\end{tabular}%
}
\caption{Comparison of different initialization methods to finetune extra background class classifier for 5-shot classification. Our model with random initialization to finetune 20 epochs for the extra background classifier has the best performance. }
\label{table:initcomparison}
\vskip -2em
\end{table}

\textbf{\emph{tiered}ImageNet}~\cite{tieredImageNet}: It is another popular FSL benchmark with more classes, which includes 608 classes from 34 super-classes. We follow ~\cite{SnaTCHer} to split training, validation, and testing into 351, 97, and 160, respectively.

\textbf{Caltech-UCSD Birds-200-2011}~\cite{cub}: CUB is a fine-grained bird classification dataset with 200 categories and in total 11,788 images. Under the 5-way FSOSR setting, we randomly sampled 5 classes as the closes set and 5 classes as the open sets from 100 meta-training classes to form the training episodes. The validation episodes and testing episodes were sampled from the rest 50 meta-validation classes and 50 meta-testing classes. 

\subsection{Implementation Details}
Following~\cite{SnaTCHer}, we use the same network backbone Resnet12 for all datasets, where the last layer before the classifier contains 640 dimensions. We built our project using PyTorch~\cite{pytorch}. The model was pre-trained with batch size of 128 for 500 epochs and optimized using SGD with momentum. The pre-training learning rate is adjusted using cosine annealing from 0.1 to 0 in a cycle of 100 epochs. Standard data augmentation with random crops, random color jitters, and random horizental flips are applied during pre-training. Manifold-mixup in conjunction with self-supervision~\cite{manifoldmixup} is also used to make backbone more robust.
Our baseline is based on the first FSOSR paper~\cite{peeler} using meta-learning strategy to finetune the backbone episodically. We use cosine similarity as our distance metric and fine-tune the network in $20{,}000$ iterations with an initial learning rate $0.0002$.
Ours ProCAM and softmax ProCAM are fine-tuning the extra background classifier with $0.002$ initial learning rate for 20 epochs. The loss weight $\lambda$ is set to 0.05 for \emph{mini}ImageNet, CUB, and 0.005 for \emph{tiered}ImageNet. We used the validation data to select the best model and evaluated the test data for 600 episodes to minimize uncertainty.
\begin{figure}[t]
	\centering
	\includegraphics[width=1\linewidth]{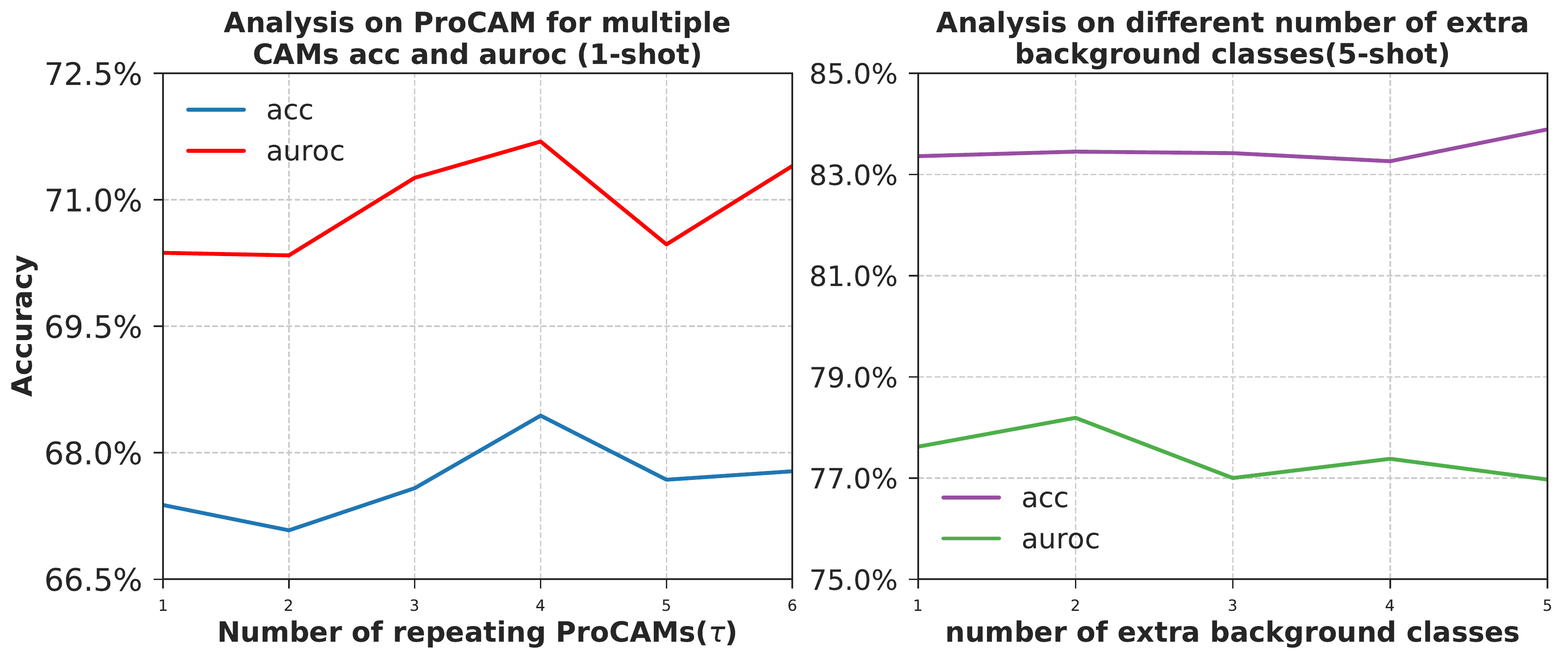}
	\caption{Analysis accuracy(\%) and auroc(\%) on different number of ProCAM and extra background class. The best number  of repeating ProCAM is $4$ times for both accuracy and AUROC on 1-shot. The highest AUROC for 5-shot is having $2$ extra background classes. Increasing the number of extra background classes does not affect the accuracy too much.  }
	\label{fig:analysi_dm}
\vskip -1em
\end{figure}

\begin{table*}[t]
	\centering
	\begin{tabularx}{0.72\textwidth}{l c | c | c | c || c | c | c | c}
	\hline
	& \multicolumn{4}{c||}{miniImageNet 5-way} & \multicolumn{4}{c}{tieredImageNet 5-way}\\
	& \multicolumn{2}{c}{1-shot} & \multicolumn{2}{c||}{5-shot} & \multicolumn{2}{c}{1-shot} & \multicolumn{2}{c}{5-shot} \\
	Model & \multicolumn{1}{c}{Acc} & \multicolumn{1}{c}{AUROC} & \multicolumn{1}{c}{Acc} & \multicolumn{1}{c||}{AUROC} & \multicolumn{1}{c}{Acc} & \multicolumn{1}{c}{AUROC} & \multicolumn{1}{c}{Acc} & \multicolumn{1}{c}{AUROC} \\
	\hline
	ProtoNet~\cite{ProtoNet} & $64.01$ & $51.81$ & $80.09$ & $60.39$ & $68.26$ & $60.73$ & $83.40$ & $64.96$ \\
	FEAT~\cite{feat} & $67.02$ & $57.01$ & $82.02$ & $63.18$ & $70.52$ & $63.54$ & $84.74$ & $70.74$ \\
	ASY~\cite{easy} & $61.64$ & $66.43$ & $82.37$ & $74.45$ & $66.28$ & $71.06$ & $82.53$ & $78.54$ \\
	\hline
	NN~\cite{NN} & $63.82$ & $56.96$ & $80.12$ & $63.43$ & $67.73$ & $62.70$ & $83.43$ & $69.77$ \\
	OpenMax~\cite{openmax} & $63.69$ & $62.64$ & $80.56$ & $62.27$ & $68.28$ & $60.13$ & $83.48$ & $65.51$ \\
	\hline
	baseline & $63.18$ & $70.22$ & $80.09$ & $75.97$ & $68.48$ & $72.98$ & $83.52$ & $80.07$ \\
	PEELER*~\cite{peeler} & $58.31$ & $61.66$ & $75.08$ & $69.85$ & $-$ & $-$ & $-$ & $-$ \\
	PEELER~\cite{peeler} & $65.86$ & $60.57$ & $80.61$ & $67.35$ & $69.51$ & $65.20$ & $84.10$ & $73.27$ \\
	SnaTCHer~\cite{SnaTCHer} & $67.02$ & $68.27$ & $82.02$ & $77.42$ & $\textbf{70.52}$ & $74.28$ & $84.74$ & $82.02$ \\

	\hline
	Ours\_ProCAM & $67.78$ & $\textbf{71.41}$ & $83.45$ & $\textbf{78.19}$ & $68.16$ & $75.35$ & $\textbf{86.34}$ & $\textbf{82.76}$ \\
	Ours\_ProCAMsm & $\textbf{67.86}$ & $71.09$ & $\textbf{83.66}$ & $77.51$ & $68.82$ & $\textbf{75.55}$ & $85.64$ & $\textbf{82.77}$ \\
	\hline
	\end{tabularx}
	\caption{Comparison with the State-of-the-art methods. Average closed-set classification accuracies (\%) and average unknown detection AUROCs (\%) over 600 episodes for MiniImageNet and TieredImageNet. PEELER* is quoted from the paper, which has a ResNet-18 backbone.}
	\label{table:sota}
\vskip -2em
\end{table*}

\begin{table}[t]
% \small
	\centering
	\begin{tabularx}{0.42\textwidth}{l c | c | c | c }
	\hline
	& \multicolumn{4}{c}{CUB 5-way} \\
	& \multicolumn{2}{c}{1-shot} & \multicolumn{2}{c}{5-shot} \\
	\hline
	Model & \multicolumn{1}{c|}{Acc} & \multicolumn{1}{c|}{AUROC} & \multicolumn{1}{c|}{Acc} & \multicolumn{1}{c}{AUROC} \\
	\hline
	baseline & $61.24$ & $68.43$ & $76.23$ & $75.09$ \\
	ProtoNet~\cite{ProtoNet} & $57.31$ & $60.32$ & $73.19$ & $64.55$  \\

	PEELER~\cite{peeler} & $59.42$ & $58.63$ & $78.42$ & $66.04$ \\
	SnaTCHer~\cite{SnaTCHer} & $57.98$ & $64.55$ & $77.05$ & $71.05$ \\

	\hline
	Ours\_ProCAM & $65.88$ & $75.88$ & $81.14$ & $\textbf{83.70}$ \\
	Ours\_ProCAMsm & $\textbf{68.54}$ & $\textbf{76.05}$ & $\textbf{82.22}$ & $82.34$ \\
	\hline
	\end{tabularx}
	\caption{Comparison with the State-of-the-art methods. Average closed-set classification accuracies (\%) and average unknown detection AUROCs (\%) over 600 episodes for CUB.}
	\label{table:comparison}
\vskip -1 em
\end{table}

\textbf{Evaluation Protocol.}
We follow ~\cite{peeler,SnaTCHer} to evaluate the model in two metrics, the known class classification accuracy (acc) and the area under the ROC curve (AUROC) for the unknown class detection. The known class accuracy takes the known query ground truth label and checks the model corrected classification ratio. The AUROC checks the area under the ROC curve. The ROC curve is plotted based on the different thresholds for the true positive rate versus the false positive rate. The higher AUROC indicates the better unseen detection capability. 
\subsection{Analysis}
We evaluate the effectiveness of different initialization methods for the background class in Table.~\ref{table:initcomparison}. For each initialization method, we compare the different number of fine-tune epochs for the background class classifier learning. The random initialization method uses the Kaiming uniform initialization~\cite{he2015delving}. The 'avg' method initializes the background classifier weight with the average background feature maps. The 'global' method uses a global FC for background class across all the training and testing episodes. The results show that random initialization for every episode and fine-tuning with 20 epochs have a better result compared to the other methods. 
Fig.~\ref{fig:analysi_dm} shows the accuracy and AUROC change w.r.t. the number of repeating ProCAMs and the number of additional background classes. There is an increasing trend for accuracy and AUROC when the number of repetitions increases and reaches a maximum at $\tau = 4$. Increasing the number of extra background classes does not affect the accuracy too much, but the AUROC changes a lot. The highest AUROC for \emph{mini}ImageNet 5-shot is adding 2 extra background classes.

\begin{figure}[th]
	\centering
	\includegraphics[width=1\linewidth]{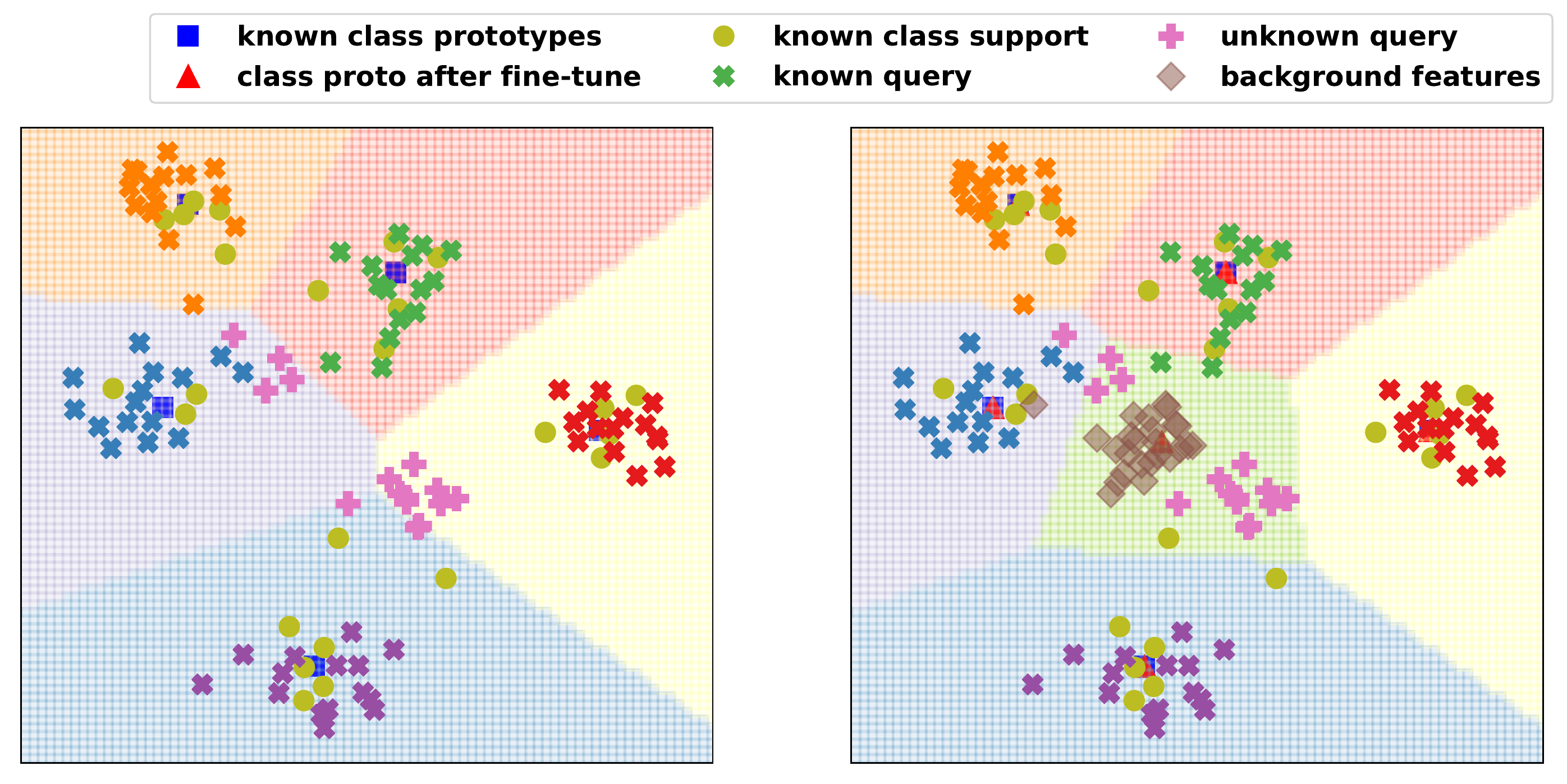}
	\caption{T-SNE~\cite{maaten2008visualizing} visualization of data embeddings and classifiers before (left) and after (right) fine-tune with background features as unknowns.
	Crosses with different colors represent query data points from different known classes. The pink plus represents unknown query. The blue squares and triangles indicate the classifier prototypes before and after fine-tuning, respectively. Our module reserve the space for the background classifier as shown in the green area.}
	\label{fig:tsne}
\vskip -1em
\end{figure}

We also plot the t-SNE figure ~\cite{maaten2008visualizing} to see the update of the feature space for the background class. Fig.~\ref{fig:tsne} shows that our proposed background classifier helps reserve space for unknown samples and provides a direction for the unknown class to cluster to the extra background class centers.

\begin{figure*}[th]
	\centering
	\includegraphics[width=0.75\linewidth]{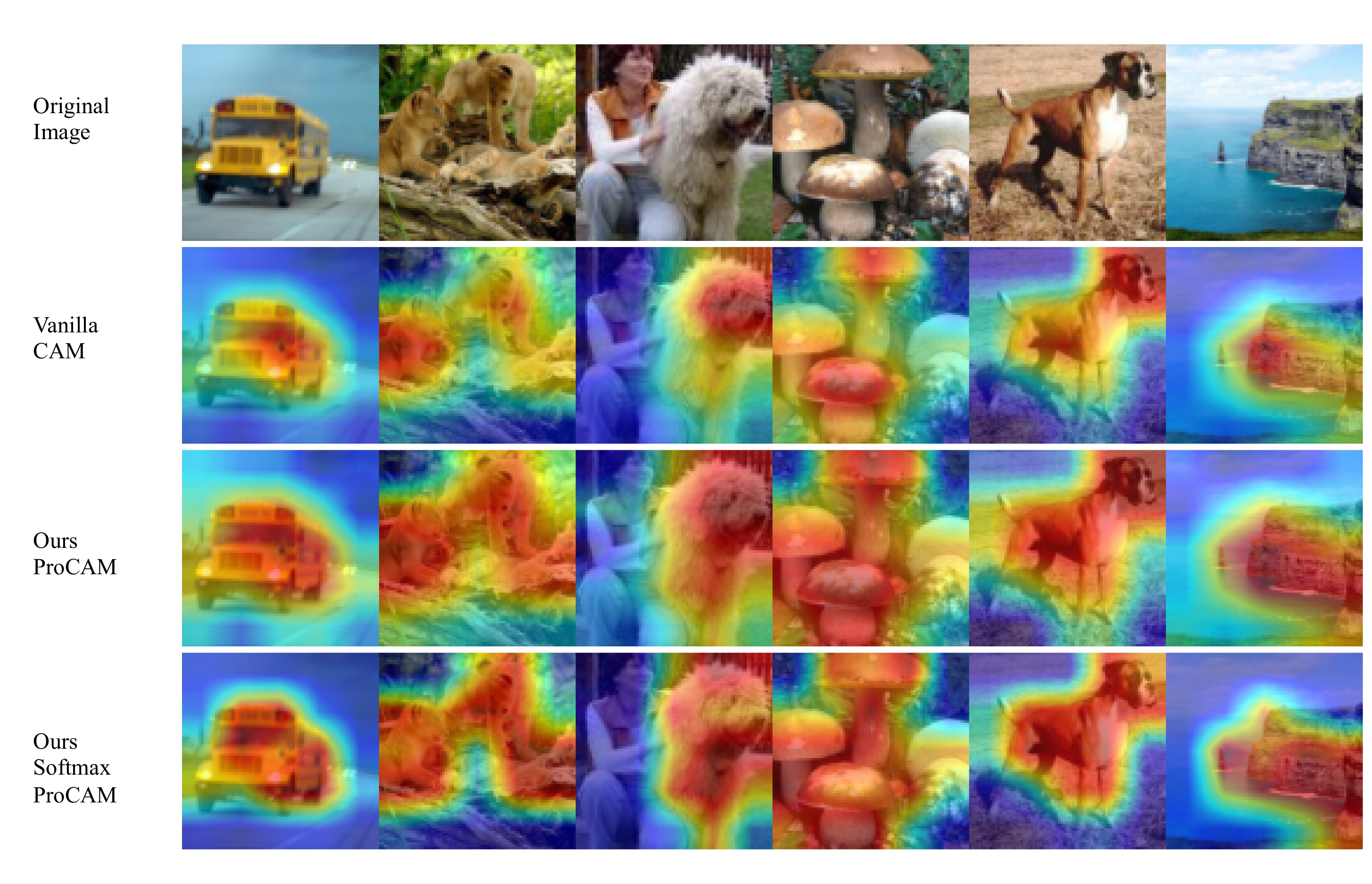}
	\vskip -2em
	\caption{Visualization of our ProCAM compare to the vanilla CAM~\cite{zhou2016learning}. Ours can generate more accurate foreground masks and get better background features that contain less information from the known classes. Our Softmax ProCAM generates better background features through the well-defined boundaries (The background regions heatmap is darker compared to the others). }
	\vskip -1em
	\label{fig:visucam}
\end{figure*}
We provide a visualization of our ProCAM and softmax ProCAM, and compared them with the original CAM in Fig.~\ref{fig:visucam}. Our methods generate more accurate foreground masks, which cover all the important features in the known classes. The background feature based on our mask contains less discriminative information for seen class and can provide good background data to train our extra background classes. The background feature data is also helpful for pushing the boundary of known classes. Softmax ProCAM is similar to the ProCAM results, the background and foreground boundaries are more clear than ProCAM.

\textbf{Ablation study.}
We perform our ablation study on the 5-way 5-shot setting for \emph{tiered}ImageNet to analyze the effectiveness of different components in~\ref{table:ablac}. The base method is the testing result for the pre-train model under FSOSR setting directly. The method 'meta\_cos' employs the meta-learning and cosine distance and gets a better result than the pre-training model. We apply the vanilla class activation map to the previous 'meta\_cos' method and we can get higher accuracy with better AUROC for this 'meta\_cam' model. The previous CAM method only uses the background feature to fine-tune the background class classifiers, we try to fine-tune all the FCs using both the support feature and the background feature in 'meta\_cam\_ft'. The result provides around 0.6\% of the increase in the accuracy, which indicates that the background features can help push the boundaries of the known classes and train better background class classifiers. We optimize the fine-tune of all FCs with better background feature maps. The results show an increasing trend in known class accuracy and unknown class AUROC.

\subsection{Comparison with the State-of-the-Art Methods}
Finally, we conduct comparison of our performance with the state-of-the-art results on three benchmarks: miniImageNet, tieredImageNet, and CUB. We choose the state-of-the-art FSL methods ProtoNet~\cite{ProtoNet}, FEAT~\cite{feat}, and ASY~\cite{easy} as the first group of models. In the second group, we select two OSR methods (NN~\cite{NN}, OpenMax~\cite{openmax}). The third group is our baseline with mete-learning and cosine classifier and the current existing FSOSR methods PEELER and SnaTCHer~\cite{peeler,SnaTCHer}. Because the existing FSL and OSR methods are not designed for FSOSR tasks, we follow SnaTCHer~\cite{SnaTCHer} to slightly modify them for comparison. 

We show the detailed numbers for miniImageNet and tieredImageNet in Table~\ref{table:sota}. The comparison for CUB is in Table~\ref{table:comparison}.
Our model has the highest AUROC of all the methods. The accuracy of the known classes is also competitive among all the datasets.
 In particular, our AUROC outperforms the state-of-the-art result by 3.14\% on \emph{mini}ImageNet 5-way 1-shot. Our accuracy increased 1.64\% for the same dataset 5-shot. For \emph{tiered}Imagenet, the 1-shot accuracy is slightly lower maybe due to the lack of support background data, but the AUROC still improved around 1.27\%. The accuracy and AUROC on CUB outperform the state-of-the-art method by 10.56\% and 11.5\%, respectively.

\section{Conclusion}

In this paper, we solve the few-shot open-set recognition problems from two aspects. We first  
introduced extra background class classifiers to reserve space for unknown samples, which use cosine distance to determine the space for each class. Then, we propose a learning paradigm to utilize the background features as unknowns to train the extra background class classifiers and fine-tune the known classes' FCs. To generate a better mask for the foreground objects and get a better background feature, we propose a ProCAM module to utilize multiple class activation maps. We optimize our ProCAM module using softmax to separate the foreground part and generate clear boundaries of the background.
Experiments on three datasets show that our method significantly outperforms the baselines and the state-of-the-art approaches.

% \section{Introduction}

% \section{SIGCHI Extended Abstracts}

% The ``\verb|sigchi-a|'' template style (available only in \LaTeX\ and
% not in Word) produces a landscape-orientation formatted article, with
% a wide left margin. Three environments are available for use with the
% ``\verb|sigchi-a|'' template style, and produce formatted output in
% the margin:
% \begin{itemize}
% \item {\verb|sidebar|}:  Place formatted text in the margin.
% \item {\verb|marginfigure|}: Place a figure in the margin.
% \item {\verb|margintable|}: Place a table in the margin.
% \end{itemize}

% %%
% %% The acknowledgments section is defined using the "acks" environment
% %% (and NOT an unnumbered section). This ensures the proper
% %% identification of the section in the article metadata, and the
% %% consistent spelling of the heading.
\begin{acks}
This work is supported by Alibaba Group through Alibaba Innovative Research (AIR) Program and Alibaba-NTU Singaproe Joint Research Institute (JRI), Nanyang Technological University, Singapore. This research is also supported by the National Research Foundation, Singapore under its AI Singapore Programme (AISG Award No: AISG-RP-2018-003), the Ministry of Education, Singapore, under its Academic Research Fund Tier 2 (MOE-T2EP20220-0007) and Tier 1 (RG95/20).
\end{acks}

\bibliographystyle{ACM-Reference-Format}
\bibliography{egbib}


\begin{thebibliography}{58}

%%% the Web address.

\ifx \showCODEN    \undefined \def \showCODEN     #1{\unskip}     \fi
\ifx \showDOI      \undefined \def \showDOI       #1{#1}\fi
\ifx \showISBNx    \undefined \def \showISBNx     #1{\unskip}     \fi
\ifx \showISBNxiii \undefined \def \showISBNxiii  #1{\unskip}     \fi
\ifx \showISSN     \undefined \def \showISSN      #1{\unskip}     \fi
\ifx \showLCCN     \undefined \def \showLCCN      #1{\unskip}     \fi
\ifx \shownote     \undefined \def \shownote      #1{#1}          \fi
\ifx \showarticletitle \undefined \def \showarticletitle #1{#1}   \fi
\ifx \showURL      \undefined \def \showURL       {\relax}        \fi
% invisible to TeX
\providecommand\bibfield[2]{#2}
\providecommand\bibinfo[2]{#2}
\providecommand\natexlab[1]{#1}
\providecommand\showeprint[2][]{arXiv:#2}

\bibitem[Arjovsky et~al\mbox{.}(2017)]%
        {wgan}
\bibfield{author}{\bibinfo{person}{Martin Arjovsky}, \bibinfo{person}{Soumith
  Chintala}, {and} \bibinfo{person}{L{\'e}on Bottou}.}
  \bibinfo{year}{2017}\natexlab{}.
\newblock \showarticletitle{Wasserstein generative adversarial networks}. In
  \bibinfo{booktitle}{\emph{Proceedings of the 34th International Conference on
  Machine Learning}}. \bibinfo{pages}{214--223}.
\newblock


\bibitem[Bendale and Boult(2016)]%
        {openmax}
\bibfield{author}{\bibinfo{person}{Abhijit Bendale} {and}
  \bibinfo{person}{Terrance~E Boult}.} \bibinfo{year}{2016}\natexlab{}.
\newblock \showarticletitle{Towards open set deep networks}. In
  \bibinfo{booktitle}{\emph{Proceedings of the IEEE conference on computer
  vision and pattern recognition}}. \bibinfo{pages}{1563--1572}.
\newblock


\bibitem[Bendou et~al\mbox{.}(2022)]%
        {easy}
\bibfield{author}{\bibinfo{person}{Yassir Bendou}, \bibinfo{person}{Yuqing Hu},
  \bibinfo{person}{Raphael Lafargue}, \bibinfo{person}{Giulia Lioi},
  \bibinfo{person}{Bastien Pasdeloup}, \bibinfo{person}{St{\'e}phane Pateux},
  {and} \bibinfo{person}{Vincent Gripon}.} \bibinfo{year}{2022}\natexlab{}.
\newblock \showarticletitle{EASY: Ensemble Augmented-Shot Y-shaped Learning:
  State-Of-The-Art Few-Shot Classification with Simple Ingredients}.
\newblock \bibinfo{journal}{\emph{arXiv preprint arXiv:2201.09699}}
  (\bibinfo{year}{2022}).
\newblock


\bibitem[Bronskill et~al\mbox{.}(2020)]%
        {tasknorm}
\bibfield{author}{\bibinfo{person}{John Bronskill}, \bibinfo{person}{Jonathan
  Gordon}, \bibinfo{person}{James Requeima}, \bibinfo{person}{Sebastian
  Nowozin}, {and} \bibinfo{person}{Richard~E Turner}.}
  \bibinfo{year}{2020}\natexlab{}.
\newblock \showarticletitle{TaskNorm: Rethinking Batch Normalization for
  Meta-Learning}.
\newblock \bibinfo{journal}{\emph{arXiv preprint arXiv:2003.03284}}
  (\bibinfo{year}{2020}).
\newblock


\bibitem[Chen et~al\mbox{.}(2020)]%
        {chen2020compositional}
\bibfield{author}{\bibinfo{person}{Xiaoyu Chen}, \bibinfo{person}{Chi Zhang},
  \bibinfo{person}{Guosheng Lin}, {and} \bibinfo{person}{Jing Han}.}
  \bibinfo{year}{2020}\natexlab{}.
\newblock \showarticletitle{Compositional Prototype Network with Multi-view
  Comparision for Few-Shot Point Cloud Semantic Segmentation}.
\newblock \bibinfo{journal}{\emph{arXiv preprint arXiv:2012.14255}}
  (\bibinfo{year}{2020}).
\newblock


\bibitem[Daihong et~al\mbox{.}(2022)]%
        {daihong2022multi}
\bibfield{author}{\bibinfo{person}{Jiang Daihong}, \bibinfo{person}{Zhang Sai},
  \bibinfo{person}{Dai Lei}, {and} \bibinfo{person}{Dai Yueming}.}
  \bibinfo{year}{2022}\natexlab{}.
\newblock \showarticletitle{Multi-scale generative adversarial network for
  image super-resolution}.
\newblock \bibinfo{journal}{\emph{Soft Computing}} \bibinfo{volume}{26},
  \bibinfo{number}{8} (\bibinfo{year}{2022}), \bibinfo{pages}{3631--3641}.
\newblock


\bibitem[Finn et~al\mbox{.}(2017)]%
        {MAML}
\bibfield{author}{\bibinfo{person}{Chelsea Finn}, \bibinfo{person}{Pieter
  Abbeel}, {and} \bibinfo{person}{Sergey Levine}.}
  \bibinfo{year}{2017}\natexlab{}.
\newblock \showarticletitle{Model-agnostic meta-learning for fast adaptation of
  deep networks}. In \bibinfo{booktitle}{\emph{Proceedings of the 34th
  International Conference on Machine Learning}}. JMLR. org,
  \bibinfo{pages}{1126--1135}.
\newblock


\bibitem[Ge et~al\mbox{.}(2017)]%
        {Gopenmax}
\bibfield{author}{\bibinfo{person}{Zongyuan Ge}, \bibinfo{person}{Sergey
  Demyanov}, \bibinfo{person}{Zetao Chen}, {and} \bibinfo{person}{Rahil
  Garnavi}.} \bibinfo{year}{2017}\natexlab{}.
\newblock \showarticletitle{Generative OpenMax for multi-class open set
  classification}. In \bibinfo{booktitle}{\emph{British Machine Vision
  Conference 2017}}. British Machine Vision Association and Society for Pattern
  Recognition.
\newblock


\bibitem[Han et~al\mbox{.}(2021)]%
        {ceGZSL}
\bibfield{author}{\bibinfo{person}{Zongyan Han}, \bibinfo{person}{Zhenyong Fu},
  \bibinfo{person}{Shuo Chen}, {and} \bibinfo{person}{Jian Yang}.}
  \bibinfo{year}{2021}\natexlab{}.
\newblock \showarticletitle{Contrastive Embedding for Generalized Zero-Shot
  Learning}. In \bibinfo{booktitle}{\emph{Proceedings of the IEEE/CVF
  Conference on Computer Vision and Pattern Recognition}}.
  \bibinfo{pages}{2371--2381}.
\newblock


\bibitem[He et~al\mbox{.}(2015)]%
        {he2015delving}
\bibfield{author}{\bibinfo{person}{Kaiming He}, \bibinfo{person}{Xiangyu
  Zhang}, \bibinfo{person}{Shaoqing Ren}, {and} \bibinfo{person}{Jian Sun}.}
  \bibinfo{year}{2015}\natexlab{}.
\newblock \showarticletitle{Delving deep into rectifiers: Surpassing
  human-level performance on imagenet classification}. In
  \bibinfo{booktitle}{\emph{Proceedings of the IEEE international conference on
  computer vision}}. \bibinfo{pages}{1026--1034}.
\newblock


\bibitem[He et~al\mbox{.}(2016)]%
        {resnet}
\bibfield{author}{\bibinfo{person}{Kaiming He}, \bibinfo{person}{Xiangyu
  Zhang}, \bibinfo{person}{Shaoqing Ren}, {and} \bibinfo{person}{Jian Sun}.}
  \bibinfo{year}{2016}\natexlab{}.
\newblock \showarticletitle{Deep residual learning for image recognition}. In
  \bibinfo{booktitle}{\emph{Proceedings of the IEEE conference on computer
  vision and pattern recognition}}. \bibinfo{pages}{770--778}.
\newblock


\bibitem[Hou et~al\mbox{.}(2019)]%
        {Canfsl}
\bibfield{author}{\bibinfo{person}{Ruibing Hou}, \bibinfo{person}{Hong Chang},
  \bibinfo{person}{MA Bingpeng}, \bibinfo{person}{Shiguang Shan}, {and}
  \bibinfo{person}{Xilin Chen}.} \bibinfo{year}{2019}\natexlab{}.
\newblock \showarticletitle{Cross attention network for few-shot
  classification}. In \bibinfo{booktitle}{\emph{Advances in Neural Information
  Processing Systems}}. \bibinfo{pages}{4003--4014}.
\newblock


\bibitem[Huang et~al\mbox{.}(2019)]%
        {densenet}
\bibfield{author}{\bibinfo{person}{Gao Huang}, \bibinfo{person}{Zhuang Liu},
  \bibinfo{person}{Geoff Pleiss}, \bibinfo{person}{Laurens Van Der~Maaten},
  {and} \bibinfo{person}{Kilian Weinberger}.} \bibinfo{year}{2019}\natexlab{}.
\newblock \showarticletitle{Convolutional Networks with Dense Connectivity}.
\newblock \bibinfo{journal}{\emph{IEEE Transactions on Pattern Analysis and
  Machine Intelligence}} (\bibinfo{year}{2019}).
\newblock


\bibitem[Jeong et~al\mbox{.}(2021)]%
        {SnaTCHer}
\bibfield{author}{\bibinfo{person}{Minki Jeong}, \bibinfo{person}{Seokeon
  Choi}, {and} \bibinfo{person}{Changick Kim}.}
  \bibinfo{year}{2021}\natexlab{}.
\newblock \showarticletitle{Few-Shot Open-Set Recognition by Transformation
  Consistency}. In \bibinfo{booktitle}{\emph{Proceedings of the IEEE/CVF
  Conference on Computer Vision and Pattern Recognition}}.
  \bibinfo{pages}{12566--12575}.
\newblock


\bibitem[J{\'u}nior et~al\mbox{.}(2017)]%
        {NN}
\bibfield{author}{\bibinfo{person}{Pedro R~Mendes J{\'u}nior},
  \bibinfo{person}{Roberto~M De~Souza}, \bibinfo{person}{Rafael de~O Werneck},
  \bibinfo{person}{Bernardo~V Stein}, \bibinfo{person}{Daniel~V Pazinato},
  \bibinfo{person}{Waldir~R de Almeida}, \bibinfo{person}{Ot{\'a}vio~AB
  Penatti}, \bibinfo{person}{Ricardo da~S Torres}, {and}
  \bibinfo{person}{Anderson Rocha}.} \bibinfo{year}{2017}\natexlab{}.
\newblock \showarticletitle{Nearest neighbors distance ratio open-set
  classifier}.
\newblock \bibinfo{journal}{\emph{Machine Learning}} \bibinfo{volume}{106},
  \bibinfo{number}{3} (\bibinfo{year}{2017}), \bibinfo{pages}{359--386}.
\newblock


\bibitem[Kong and Ramanan(2021)]%
        {OpenGAN}
\bibfield{author}{\bibinfo{person}{Shu Kong} {and} \bibinfo{person}{Deva
  Ramanan}.} \bibinfo{year}{2021}\natexlab{}.
\newblock \showarticletitle{OpenGAN: Open-Set Recognition via Open Data
  Generation}. In \bibinfo{booktitle}{\emph{Proceedings of the IEEE
  International Conference on Computer Vision}}.
\newblock


\bibitem[Krizhevsky et~al\mbox{.}(2012)]%
        {Alexnet}
\bibfield{author}{\bibinfo{person}{Alex Krizhevsky}, \bibinfo{person}{Ilya
  Sutskever}, {and} \bibinfo{person}{Geoffrey~E. Hinton}.}
  \bibinfo{year}{2012}\natexlab{}.
\newblock \showarticletitle{ImageNet Classification with Deep Convolutional
  Neural Networks}.
\newblock In \bibinfo{booktitle}{\emph{Advances in Neural Information
  Processing Systems 25}}, \bibfield{editor}{\bibinfo{person}{F.~Pereira},
  \bibinfo{person}{C.~J.~C. Burges}, \bibinfo{person}{L.~Bottou}, {and}
  \bibinfo{person}{K.~Q. Weinberger}} (Eds.). \bibinfo{publisher}{Curran
  Associates, Inc.}, \bibinfo{pages}{1097--1105}.
\newblock


\bibitem[Lee et~al\mbox{.}(2019)]%
        {lee2019meta}
\bibfield{author}{\bibinfo{person}{Hae~Beom Lee}, \bibinfo{person}{Taewook
  Nam}, \bibinfo{person}{Eunho Yang}, {and} \bibinfo{person}{Sung~Ju Hwang}.}
  \bibinfo{year}{2019}\natexlab{}.
\newblock \showarticletitle{Meta Dropout: Learning to Perturb Latent Features
  for Generalization}. In \bibinfo{booktitle}{\emph{International Conference on
  Learning Representations}}.
\newblock


\bibitem[Li et~al\mbox{.}(2019)]%
        {CTM}
\bibfield{author}{\bibinfo{person}{Hongyang Li}, \bibinfo{person}{David Eigen},
  \bibinfo{person}{Samuel Dodge}, \bibinfo{person}{Matthew Zeiler}, {and}
  \bibinfo{person}{Xiaogang Wang}.} \bibinfo{year}{2019}\natexlab{}.
\newblock \showarticletitle{Finding task-relevant features for few-shot
  learning by category traversal}. In \bibinfo{booktitle}{\emph{Proceedings of
  the IEEE Conference on Computer Vision and Pattern Recognition}}.
  \bibinfo{pages}{1--10}.
\newblock


\bibitem[Li et~al\mbox{.}(2022)]%
        {li2022rigidflow}
\bibfield{author}{\bibinfo{person}{Ruibo Li}, \bibinfo{person}{Chi Zhang},
  \bibinfo{person}{Guosheng Lin}, \bibinfo{person}{Zhe Wang}, {and}
  \bibinfo{person}{Chunhua Shen}.} \bibinfo{year}{2022}\natexlab{}.
\newblock \showarticletitle{RigidFlow: Self-Supervised Scene Flow Learning on
  Point Clouds by Local Rigidity Prior}. In
  \bibinfo{booktitle}{\emph{Proceedings of the IEEE/CVF Conference on Computer
  Vision and Pattern Recognition}}. \bibinfo{pages}{16959--16968}.
\newblock


\bibitem[Liu et~al\mbox{.}(2020a)]%
        {peeler}
\bibfield{author}{\bibinfo{person}{Bo Liu}, \bibinfo{person}{Hao Kang},
  \bibinfo{person}{Haoxiang Li}, \bibinfo{person}{Gang Hua}, {and}
  \bibinfo{person}{Nuno Vasconcelos}.} \bibinfo{year}{2020}\natexlab{a}.
\newblock \showarticletitle{Few-Shot Open-Set Recognition using Meta-Learning}.
  In \bibinfo{booktitle}{\emph{Proceedings of the IEEE/CVF Conference on
  Computer Vision and Pattern Recognition}}. \bibinfo{pages}{8798--8807}.
\newblock


\bibitem[Liu et~al\mbox{.}(2018)]%
        {liu2018generalized}
\bibfield{author}{\bibinfo{person}{Shichen Liu}, \bibinfo{person}{Mingsheng
  Long}, \bibinfo{person}{Jianmin Wang}, {and} \bibinfo{person}{Michael~I
  Jordan}.} \bibinfo{year}{2018}\natexlab{}.
\newblock \showarticletitle{Generalized zero-shot learning with deep
  calibration network}. In \bibinfo{booktitle}{\emph{Advances in Neural
  Information Processing Systems}}. \bibinfo{pages}{2005--2015}.
\newblock


\bibitem[Liu et~al\mbox{.}(2021)]%
        {liu2021fewshot}
\bibfield{author}{\bibinfo{person}{Weide Liu}, \bibinfo{person}{Chi Zhang},
  \bibinfo{person}{Henghui Ding}, \bibinfo{person}{Tzu-Yi Hung}, {and}
  \bibinfo{person}{Guosheng Lin}.} \bibinfo{year}{2021}\natexlab{}.
\newblock \bibinfo{title}{Few-shot Segmentation with Optimal Transport Matching
  and Message Flow}.
\newblock
\newblock
\showeprint[arxiv]{2108.08518}~[cs.CV]


\bibitem[Liu et~al\mbox{.}(2020c)]%
        {liu2020weakly}
\bibfield{author}{\bibinfo{person}{Weide Liu}, \bibinfo{person}{Chi Zhang},
  \bibinfo{person}{Guosheng Lin}, \bibinfo{person}{Tzu-Yi HUNG}, {and}
  \bibinfo{person}{Chunyan Miao}.} \bibinfo{year}{2020}\natexlab{c}.
\newblock \showarticletitle{Weakly Supervised Segmentation with Maximum
  Bipartite Graph Matching}. In \bibinfo{booktitle}{\emph{Proceedings of the
  28th ACM International Conference on Multimedia}}.
  \bibinfo{pages}{2085--2094}.
\newblock


\bibitem[Liu et~al\mbox{.}(2020b)]%
        {liu2020crnet}
\bibfield{author}{\bibinfo{person}{Weide Liu}, \bibinfo{person}{Chi Zhang},
  \bibinfo{person}{Guosheng Lin}, {and} \bibinfo{person}{Fayao Liu}.}
  \bibinfo{year}{2020}\natexlab{b}.
\newblock \showarticletitle{CRNet: Cross-Reference Networks for Few-Shot
  Segmentation}. In \bibinfo{booktitle}{\emph{Proceedings of the IEEE/CVF
  Conference on Computer Vision and Pattern Recognition}}.
  \bibinfo{pages}{4165--4173}.
\newblock


\bibitem[Maaten and Hinton(2008)]%
        {maaten2008visualizing}
\bibfield{author}{\bibinfo{person}{Laurens van~der Maaten} {and}
  \bibinfo{person}{Geoffrey Hinton}.} \bibinfo{year}{2008}\natexlab{}.
\newblock \showarticletitle{Visualizing data using t-SNE}.
\newblock \bibinfo{journal}{\emph{Journal of machine learning research}}
  \bibinfo{volume}{9}, \bibinfo{number}{Nov} (\bibinfo{year}{2008}),
  \bibinfo{pages}{2579--2605}.
\newblock


\bibitem[Mangla et~al\mbox{.}(2020)]%
        {manifoldmixup}
\bibfield{author}{\bibinfo{person}{Puneet Mangla}, \bibinfo{person}{Nupur
  Kumari}, \bibinfo{person}{Abhishek Sinha}, \bibinfo{person}{Mayank Singh},
  \bibinfo{person}{Balaji Krishnamurthy}, {and} \bibinfo{person}{Vineeth~N
  Balasubramanian}.} \bibinfo{year}{2020}\natexlab{}.
\newblock \showarticletitle{Charting the right manifold: Manifold mixup for
  few-shot learning}. In \bibinfo{booktitle}{\emph{Proceedings of the IEEE/CVF
  Winter Conference on Applications of Computer Vision}}.
  \bibinfo{pages}{2218--2227}.
\newblock


\bibitem[Neal et~al\mbox{.}(2018)]%
        {counterfactual}
\bibfield{author}{\bibinfo{person}{Lawrence Neal}, \bibinfo{person}{Matthew
  Olson}, \bibinfo{person}{Xiaoli Fern}, \bibinfo{person}{Weng-Keen Wong},
  {and} \bibinfo{person}{Fuxin Li}.} \bibinfo{year}{2018}\natexlab{}.
\newblock \showarticletitle{Open set learning with counterfactual images}. In
  \bibinfo{booktitle}{\emph{Proceedings of the European Conference on Computer
  Vision}}. \bibinfo{pages}{613--628}.
\newblock


\bibitem[Oza and Patel(2019)]%
        {c2ae}
\bibfield{author}{\bibinfo{person}{Poojan Oza} {and} \bibinfo{person}{Vishal~M
  Patel}.} \bibinfo{year}{2019}\natexlab{}.
\newblock \showarticletitle{C2ae: Class conditioned auto-encoder for open-set
  recognition}. In \bibinfo{booktitle}{\emph{Proceedings of the IEEE Conference
  on Computer Vision and Pattern Recognition}}. \bibinfo{pages}{2307--2316}.
\newblock


\bibitem[Paszke et~al\mbox{.}(2019)]%
        {pytorch}
\bibfield{author}{\bibinfo{person}{Adam Paszke}, \bibinfo{person}{Sam Gross},
  \bibinfo{person}{Francisco Massa}, \bibinfo{person}{Adam Lerer},
  \bibinfo{person}{James Bradbury}, \bibinfo{person}{Gregory Chanan},
  \bibinfo{person}{Trevor Killeen}, \bibinfo{person}{Zeming Lin},
  \bibinfo{person}{Natalia Gimelshein}, \bibinfo{person}{Luca Antiga},
  \bibinfo{person}{Alban Desmaison}, \bibinfo{person}{Andreas Kopf},
  \bibinfo{person}{Edward Yang}, \bibinfo{person}{Zachary DeVito},
  \bibinfo{person}{Martin Raison}, \bibinfo{person}{Alykhan Tejani},
  \bibinfo{person}{Sasank Chilamkurthy}, \bibinfo{person}{Benoit Steiner},
  \bibinfo{person}{Lu Fang}, \bibinfo{person}{Junjie Bai}, {and}
  \bibinfo{person}{Soumith Chintala}.} \bibinfo{year}{2019}\natexlab{}.
\newblock \showarticletitle{PyTorch: An Imperative Style, High-Performance Deep
  Learning Library}.
\newblock In \bibinfo{booktitle}{\emph{Advances in Neural Information
  Processing Systems 32}}. \bibinfo{publisher}{Curran Associates, Inc.},
  \bibinfo{pages}{8024--8035}.
\newblock


\bibitem[Perera et~al\mbox{.}(2020)]%
        {perera2020generative}
\bibfield{author}{\bibinfo{person}{Pramuditha Perera}, \bibinfo{person}{Vlad~I
  Morariu}, \bibinfo{person}{Rajiv Jain}, \bibinfo{person}{Varun Manjunatha},
  \bibinfo{person}{Curtis Wigington}, \bibinfo{person}{Vicente Ordonez}, {and}
  \bibinfo{person}{Vishal~M Patel}.} \bibinfo{year}{2020}\natexlab{}.
\newblock \showarticletitle{Generative-Discriminative Feature Representations
  for Open-Set Recognition}. In \bibinfo{booktitle}{\emph{Proceedings of the
  IEEE/CVF Conference on Computer Vision and Pattern Recognition}}.
  \bibinfo{pages}{11814--11823}.
\newblock


\bibitem[Ravi and Larochelle(2017)]%
        {omfsl}
\bibfield{author}{\bibinfo{person}{Sachin Ravi} {and} \bibinfo{person}{Hugo
  Larochelle}.} \bibinfo{year}{2017}\natexlab{}.
\newblock \showarticletitle{Optimization as a model for few-shot learning}. In
  \bibinfo{booktitle}{\emph{International Conference on Learning
  Representations}}.
\newblock


\bibitem[Ren et~al\mbox{.}(2018)]%
        {tieredImageNet}
\bibfield{author}{\bibinfo{person}{Mengye Ren}, \bibinfo{person}{Sachin Ravi},
  \bibinfo{person}{Eleni Triantafillou}, \bibinfo{person}{Jake Snell},
  \bibinfo{person}{Kevin Swersky}, \bibinfo{person}{Josh~B. Tenenbaum},
  \bibinfo{person}{Hugo Larochelle}, {and} \bibinfo{person}{Richard~S. Zemel}.}
  \bibinfo{year}{2018}\natexlab{}.
\newblock \showarticletitle{Meta-Learning for Semi-Supervised Few-Shot
  Classification}. In \bibinfo{booktitle}{\emph{International Conference on
  Learning Representations}}.
\newblock


\bibitem[Rusu et~al\mbox{.}(2019)]%
        {LEO}
\bibfield{author}{\bibinfo{person}{Andrei~A. Rusu}, \bibinfo{person}{Dushyant
  Rao}, \bibinfo{person}{Jakub Sygnowski}, \bibinfo{person}{Oriol Vinyals},
  \bibinfo{person}{Razvan Pascanu}, \bibinfo{person}{Simon Osindero}, {and}
  \bibinfo{person}{Raia Hadsell}.} \bibinfo{year}{2019}\natexlab{}.
\newblock \showarticletitle{Meta-Learning with Latent Embedding Optimization}.
  In \bibinfo{booktitle}{\emph{International Conference on Learning
  Representations}}.
\newblock


\bibitem[Simonyan and Zisserman(2014)]%
        {vgg}
\bibfield{author}{\bibinfo{person}{Karen Simonyan} {and}
  \bibinfo{person}{Andrew Zisserman}.} \bibinfo{year}{2014}\natexlab{}.
\newblock \showarticletitle{Very deep convolutional networks for large-scale
  image recognition}.
\newblock \bibinfo{journal}{\emph{arXiv preprint arXiv:1409.1556}}
  (\bibinfo{year}{2014}).
\newblock


\bibitem[Snell et~al\mbox{.}(2017)]%
        {ProtoNet}
\bibfield{author}{\bibinfo{person}{Jake Snell}, \bibinfo{person}{Kevin
  Swersky}, {and} \bibinfo{person}{Richard Zemel}.}
  \bibinfo{year}{2017}\natexlab{}.
\newblock \showarticletitle{Prototypical networks for few-shot learning}. In
  \bibinfo{booktitle}{\emph{Advances in neural information processing
  systems}}. \bibinfo{pages}{4077--4087}.
\newblock


\bibitem[Sun et~al\mbox{.}(2019)]%
        {Sun_2019_CVPR}
\bibfield{author}{\bibinfo{person}{Qianru Sun}, \bibinfo{person}{Yaoyao Liu},
  \bibinfo{person}{Tat-Seng Chua}, {and} \bibinfo{person}{Bernt Schiele}.}
  \bibinfo{year}{2019}\natexlab{}.
\newblock \showarticletitle{Meta-Transfer Learning for Few-Shot Learning}. In
  \bibinfo{booktitle}{\emph{Proceedings of the IEEE/CVF Conference on Computer
  Vision and Pattern Recognition}}.
\newblock


\bibitem[Sun et~al\mbox{.}(2021)]%
        {sun2021m2iosr}
\bibfield{author}{\bibinfo{person}{Xin Sun}, \bibinfo{person}{Henghui Ding},
  \bibinfo{person}{Chi Zhang}, \bibinfo{person}{Guosheng Lin}, {and}
  \bibinfo{person}{Keck-Voon Ling}.} \bibinfo{year}{2021}\natexlab{}.
\newblock \showarticletitle{M2iosr: Maximal mutual information open set
  recognition}.
\newblock \bibinfo{journal}{\emph{arXiv preprint arXiv:2108.02373}}
  (\bibinfo{year}{2021}).
\newblock


\bibitem[Sun et~al\mbox{.}(2020a)]%
        {sun2020conditional}
\bibfield{author}{\bibinfo{person}{Xin Sun}, \bibinfo{person}{Zhenning Yang},
  \bibinfo{person}{Chi Zhang}, \bibinfo{person}{Keck-Voon Ling}, {and}
  \bibinfo{person}{Guohao Peng}.} \bibinfo{year}{2020}\natexlab{a}.
\newblock \showarticletitle{Conditional Gaussian Distribution Learning for Open
  Set Recognition}. In \bibinfo{booktitle}{\emph{Proceedings of the IEEE/CVF
  Conference on Computer Vision and Pattern Recognition}}.
  \bibinfo{pages}{13480--13489}.
\newblock


\bibitem[Sun et~al\mbox{.}(2020b)]%
        {sun2020open}
\bibfield{author}{\bibinfo{person}{Xin Sun}, \bibinfo{person}{Chi Zhang},
  \bibinfo{person}{Guosheng Lin}, {and} \bibinfo{person}{Keck-Voon Ling}.}
  \bibinfo{year}{2020}\natexlab{b}.
\newblock \showarticletitle{Open Set Recognition with Conditional Probabilistic
  Generative Models}.
\newblock \bibinfo{journal}{\emph{arXiv preprint arXiv:2008.05129}}
  (\bibinfo{year}{2020}).
\newblock


\bibitem[Sung et~al\mbox{.}(2018)]%
        {RelationNet}
\bibfield{author}{\bibinfo{person}{Flood Sung}, \bibinfo{person}{Yongxin Yang},
  \bibinfo{person}{Li Zhang}, \bibinfo{person}{Tao Xiang},
  \bibinfo{person}{Philip~HS Torr}, {and} \bibinfo{person}{Timothy~M
  Hospedales}.} \bibinfo{year}{2018}\natexlab{}.
\newblock \showarticletitle{Learning to compare: Relation network for few-shot
  learning}. In \bibinfo{booktitle}{\emph{Proceedings of the IEEE Conference on
  Computer Vision and Pattern Recognition}}. \bibinfo{pages}{1199--1208}.
\newblock


\bibitem[Verma et~al\mbox{.}(2019)]%
        {verma2019manifold}
\bibfield{author}{\bibinfo{person}{Vikas Verma}, \bibinfo{person}{Alex Lamb},
  \bibinfo{person}{Christopher Beckham}, \bibinfo{person}{Amir Najafi},
  \bibinfo{person}{Ioannis Mitliagkas}, \bibinfo{person}{David Lopez-Paz},
  {and} \bibinfo{person}{Yoshua Bengio}.} \bibinfo{year}{2019}\natexlab{}.
\newblock \showarticletitle{Manifold mixup: Better representations by
  interpolating hidden states}. In \bibinfo{booktitle}{\emph{International
  Conference on Machine Learning}}. PMLR, \bibinfo{pages}{6438--6447}.
\newblock


\bibitem[Vinyals et~al\mbox{.}(2016)]%
        {MatchingNet}
\bibfield{author}{\bibinfo{person}{Oriol Vinyals}, \bibinfo{person}{Charles
  Blundell}, \bibinfo{person}{Timothy Lillicrap}, \bibinfo{person}{Daan
  Wierstra}, {et~al\mbox{.}}} \bibinfo{year}{2016}\natexlab{}.
\newblock \showarticletitle{Matching networks for one shot learning}. In
  \bibinfo{booktitle}{\emph{Advances in neural information processing
  systems}}. \bibinfo{pages}{3630--3638}.
\newblock


\bibitem[Wah et~al\mbox{.}(2011)]%
        {cub}
\bibfield{author}{\bibinfo{person}{C. Wah}, \bibinfo{person}{S. Branson},
  \bibinfo{person}{P. Welinder}, \bibinfo{person}{P. Perona}, {and}
  \bibinfo{person}{S. Belongie}.} \bibinfo{year}{2011}\natexlab{}.
\newblock \bibinfo{booktitle}{\emph{{The Caltech-UCSD Birds-200-2011
  Dataset}}}.
\newblock \bibinfo{type}{{T}echnical {R}eport} CNS-TR-2011-001.
  \bibinfo{institution}{California Institute of Technology}.
\newblock


\bibitem[Yang et~al\mbox{.}(2022)]%
        {yang2022efficient}
\bibfield{author}{\bibinfo{person}{Ze Yang}, \bibinfo{person}{Chi Zhang},
  \bibinfo{person}{Ruibo Li}, {and} \bibinfo{person}{Guosheng Lin}.}
  \bibinfo{year}{2022}\natexlab{}.
\newblock \showarticletitle{Efficient Few-Shot Object Detection via Knowledge
  Inheritance}.
\newblock \bibinfo{journal}{\emph{arXiv preprint arXiv:2203.12224}}
  (\bibinfo{year}{2022}).
\newblock


\bibitem[Ye et~al\mbox{.}(2019)]%
        {ye2019learning}
\bibfield{author}{\bibinfo{person}{Han-Jia Ye}, \bibinfo{person}{Hexiang Hu},
  \bibinfo{person}{De-Chuan Zhan}, {and} \bibinfo{person}{Fei Sha}.}
  \bibinfo{year}{2019}\natexlab{}.
\newblock \showarticletitle{Learning adaptive classifiers synthesis for
  generalized few-shot learning}.
\newblock \bibinfo{journal}{\emph{arXiv preprint arXiv:1906.02944}}
  (\bibinfo{year}{2019}).
\newblock


\bibitem[Ye et~al\mbox{.}(2020)]%
        {feat}
\bibfield{author}{\bibinfo{person}{Han-Jia Ye}, \bibinfo{person}{Hexiang Hu},
  \bibinfo{person}{De-Chuan Zhan}, {and} \bibinfo{person}{Fei Sha}.}
  \bibinfo{year}{2020}\natexlab{}.
\newblock \showarticletitle{Few-Shot Learning via Embedding Adaptation With
  Set-to-Set Functions}. In \bibinfo{booktitle}{\emph{Proceedings of the
  IEEE/CVF Conference on Computer Vision and Pattern Recognition}}.
\newblock


\bibitem[Yoshihashi et~al\mbox{.}(2019)]%
        {CrOSR}
\bibfield{author}{\bibinfo{person}{Ryota Yoshihashi}, \bibinfo{person}{Wen
  Shao}, \bibinfo{person}{Rei Kawakami}, \bibinfo{person}{Shaodi You},
  \bibinfo{person}{Makoto Iida}, {and} \bibinfo{person}{Takeshi Naemura}.}
  \bibinfo{year}{2019}\natexlab{}.
\newblock \showarticletitle{Classification-reconstruction learning for open-set
  recognition}. In \bibinfo{booktitle}{\emph{Proceedings of the IEEE/CVF
  Conference on Computer Vision and Pattern Recognition}}.
  \bibinfo{pages}{4016--4025}.
\newblock


\bibitem[Zhang et~al\mbox{.}(2020a)]%
        {zhang2020deepemdv2}
\bibfield{author}{\bibinfo{person}{Chi Zhang}, \bibinfo{person}{Yujun Cai},
  \bibinfo{person}{Guosheng Lin}, {and} \bibinfo{person}{Chunhua Shen}.}
  \bibinfo{year}{2020}\natexlab{a}.
\newblock \showarticletitle{DeepEMD: Differentiable Earth Mover's Distance for
  Few-Shot Learning}.
\newblock \bibinfo{journal}{\emph{arXiv e-prints}} (\bibinfo{year}{2020}).
\newblock


\bibitem[Zhang et~al\mbox{.}(2020b)]%
        {Zhang_2020_CVPR}
\bibfield{author}{\bibinfo{person}{Chi Zhang}, \bibinfo{person}{Yujun Cai},
  \bibinfo{person}{Guosheng Lin}, {and} \bibinfo{person}{Chunhua Shen}.}
  \bibinfo{year}{2020}\natexlab{b}.
\newblock \showarticletitle{DeepEMD: Few-Shot Image Classification With
  Differentiable Earth Mover's Distance and Structured Classifiers}. In
  \bibinfo{booktitle}{\emph{Proceedings of the IEEE/CVF Conference on Computer
  Vision and Pattern Recognition}}.
\newblock


\bibitem[Zhang et~al\mbox{.}(2021a)]%
        {zhang2021navigator}
\bibfield{author}{\bibinfo{person}{Chi Zhang}, \bibinfo{person}{Henghui Ding},
  \bibinfo{person}{Guosheng Lin}, \bibinfo{person}{Ruibo Li},
  \bibinfo{person}{Changhu Wang}, {and} \bibinfo{person}{Chunhua Shen}.}
  \bibinfo{year}{2021}\natexlab{a}.
\newblock \showarticletitle{Meta Navigator: Search for a Good Adaptation Policy
  for Few-shot Learning}. In \bibinfo{booktitle}{\emph{IEEE International
  Conference on Computer Vision (ICCV)}}.
\newblock


\bibitem[Zhang et~al\mbox{.}(2021b)]%
        {zhang2021cyclesegnet}
\bibfield{author}{\bibinfo{person}{Chi Zhang}, \bibinfo{person}{Guankai Li},
  \bibinfo{person}{Guosheng Lin}, \bibinfo{person}{Qingyao Wu}, {and}
  \bibinfo{person}{Rui Yao}.} \bibinfo{year}{2021}\natexlab{b}.
\newblock \showarticletitle{Cyclesegnet: Object co-segmentation with cycle
  refinement and region correspondence}.
\newblock \bibinfo{journal}{\emph{IEEE Transactions on Image Processing}}
  (\bibinfo{year}{2021}).
\newblock


\bibitem[Zhang et~al\mbox{.}(2019a)]%
        {pgnet}
\bibfield{author}{\bibinfo{person}{Chi Zhang}, \bibinfo{person}{Guosheng Lin},
  \bibinfo{person}{Fayao Liu}, \bibinfo{person}{Jiushuang Guo},
  \bibinfo{person}{Qingyao Wu}, {and} \bibinfo{person}{Rui Yao}.}
  \bibinfo{year}{2019}\natexlab{a}.
\newblock \showarticletitle{Pyramid graph networks with connection attentions
  for region-based one-shot semantic segmentation}. In
  \bibinfo{booktitle}{\emph{Proceedings of the IEEE International Conference on
  Computer Vision}}. \bibinfo{pages}{9587--9595}.
\newblock


\bibitem[Zhang et~al\mbox{.}(2019b)]%
        {zhang2019canet}
\bibfield{author}{\bibinfo{person}{Chi Zhang}, \bibinfo{person}{Guosheng Lin},
  \bibinfo{person}{Fayao Liu}, \bibinfo{person}{Rui Yao}, {and}
  \bibinfo{person}{Chunhua Shen}.} \bibinfo{year}{2019}\natexlab{b}.
\newblock \showarticletitle{CANet: Class-Agnostic Segmentation Networks with
  Iterative Refinement and Attentive Few-Shot Learning}. In
  \bibinfo{booktitle}{\emph{Proceedings of the IEEE Conference on Computer
  Vision and Pattern Recognition}}. \bibinfo{pages}{5217--5226}.
\newblock


\bibitem[Zhang et~al\mbox{.}(2021c)]%
        {Zhang_2021_CVPR}
\bibfield{author}{\bibinfo{person}{Chi Zhang}, \bibinfo{person}{Nan Song},
  \bibinfo{person}{Guosheng Lin}, \bibinfo{person}{Yun Zheng},
  \bibinfo{person}{Pan Pan}, {and} \bibinfo{person}{Yinghui Xu}.}
  \bibinfo{year}{2021}\natexlab{c}.
\newblock \showarticletitle{Few-Shot Incremental Learning with Continually
  Evolved Classifiers}. In \bibinfo{booktitle}{\emph{Proceedings of the
  IEEE/CVF Conference on Computer Vision and Pattern Recognition}}.
\newblock


\bibitem[Zhang et~al\mbox{.}(2018)]%
        {zhang2018efficient}
\bibfield{author}{\bibinfo{person}{Chi Zhang}, \bibinfo{person}{Rui Yao}, {and}
  \bibinfo{person}{Jinpeng Cai}.} \bibinfo{year}{2018}\natexlab{}.
\newblock \showarticletitle{Efficient eye typing with 9-direction gaze
  estimation}.
\newblock \bibinfo{journal}{\emph{Multimedia Tools and Applications}}
  \bibinfo{volume}{77}, \bibinfo{number}{15} (\bibinfo{year}{2018}),
  \bibinfo{pages}{19679--19696}.
\newblock


\bibitem[Zhou et~al\mbox{.}(2016)]%
        {zhou2016learning}
\bibfield{author}{\bibinfo{person}{Bolei Zhou}, \bibinfo{person}{Aditya
  Khosla}, \bibinfo{person}{Agata Lapedriza}, \bibinfo{person}{Aude Oliva},
  {and} \bibinfo{person}{Antonio Torralba}.} \bibinfo{year}{2016}\natexlab{}.
\newblock \showarticletitle{Learning deep features for discriminative
  localization}. In \bibinfo{booktitle}{\emph{Proceedings of the IEEE
  conference on computer vision and pattern recognition}}.
  \bibinfo{pages}{2921--2929}.
\newblock


\bibitem[Zhou et~al\mbox{.}(2021)]%
        {zhou2021learning}
\bibfield{author}{\bibinfo{person}{Da-Wei Zhou}, \bibinfo{person}{Han-Jia Ye},
  {and} \bibinfo{person}{De-Chuan Zhan}.} \bibinfo{year}{2021}\natexlab{}.
\newblock \showarticletitle{Learning Placeholders for Open-Set Recognition}. In
  \bibinfo{booktitle}{\emph{Proceedings of the IEEE/CVF Conference on Computer
  Vision and Pattern Recognition}}. \bibinfo{pages}{4401--4410}.
\newblock


\end{thebibliography}

\appendix









\end{document}